\documentclass[submission, Phys]{SciPost}

\usepackage{graphicx}		
\usepackage{dcolumn}		
\usepackage{bm}				
\usepackage{hyperref}		
\usepackage[caption=false]{subfig}
\usepackage{amsfonts}
\usepackage{amssymb}
\usepackage{amsmath}
\usepackage{mathtools}
\usepackage{commath}
\usepackage{enumitem}
\usepackage{mleftright}

\newcount\dotcnt\newdimen\deltay
\def\Ddot#1#2(#3,#4,#5,#6){\deltay=#6\setbox1=\hbox to0pt{\smash{\dotcnt=1
\kern#3\loop\raise\dotcnt\deltay\hbox to0pt{\hss#2}\kern#5\ifnum\dotcnt<#1
\advance\dotcnt 1\repeat}\hss}\setbox2=\vtop{\box1}\ht2=#4\box2}

\DeclareMathOperator{\erf}{erf}

\begin{document}

\begin{center}{\Large \textbf{
Transmission Matrix Inference via Pseudolikelihood Decimation
}}\end{center}

\begin{center}
D. Ancora\textsuperscript{1*},
L. Leuzzi\textsuperscript{1,2},
\end{center}

\begin{center}
{\bf 1} Institute of Nanotechnology, Consiglio Nazionale delle Ricerche, CNR-NANOTEC - Soft and Living Matter Laboratory, Piazzale A. Moro 2, I-00185, Roma, Italy
\\
{\bf 2} Dipartimento di Fisica, Università di Roma “Sapienza”, Piazzale A. Moro 2, I-00185, Roma, Italy 
\\
* daniele.ancora@nanotec.cnr.it
\end{center}

\begin{center}
\today
\end{center}


\section*{Abstract}
{\bf
One of the biggest challenges in the field of biomedical imaging is the comprehension and the exploitation of the photon scattering through disordered media. Many studies have pursued the solution to this puzzle, achieving light-focusing control or reconstructing images in complex media. In the present work, we investigate how statistical inference helps the calculation of the transmission matrix in a complex scrambling environment, enabling its usage like a normal optical element. We convert a linear input-output transmission problem into a statistical formulation based on pseudolikelihood maximization, learning the coupling matrix via random sampling of intensity realizations. Our aim is to uncover insights from the scattering problem, encouraging the development of novel imaging techniques for better medical investigations, borrowing a number of statistical tools from spin-glass theory.
}


\vspace{10pt}
\noindent\rule{\textwidth}{1pt}
\tableofcontents\thispagestyle{fancy}
\noindent\rule{\textwidth}{1pt}
\vspace{10pt}

\section{Introduction}
\label{sec:intro}
Linear problems are largely used in many scientific fields, among which modern topics range from general machine learning aspects \cite{robert2014machine, friedman2001elements}, to spectra deconvolution \cite{peckner2018specter}, stock market forecast \cite{altay2005stock} till the generalization of signal transmission through disordered systems \cite{Popoff2011}. Among its generality, it is particularly in the latter that our study finds its main motivation. The non-trivial problem of the transmission matrix recovery of an opaque device would enable its usage as an ordinary optical tool, capable to focus or transmit an image through disorder \cite{bertolotti2012non, vellekoop2010exploiting}, turning the scattering into a beneficial feature exploitable to enhance resolution. Turbidity is a random feature of the media depending upon microscopic arrangement of its constituents, and several approaches have been studied to investigate imaging properties in such conditions \cite{sebbah2001waves}. The light traveling through a non-uniform environment experiences several changes in its original trajectory due to a random distribution of the refractive index, loosing the memory of its initial direction. Although the microscopic process is complicated, the overall effect can be efficiently described with the definition of a transmission matrix \cite{Popoff2011}. Such matrix describes how an input field is linearly transported into the corresponding output by any arbitrary media and contains the rules on how the medium acts on a given input \cite{Popoff2011, Yoon2015, Carpenter2014}. Defining a robust way to measure such matrix is one of the greatest challenges in the field of disordered optics \cite{Wiersma2013} and opens up the usage of new tools for biomedical inspections, such as multimode-fibers \cite{Flaes2018}. It is not the interest of this paper to describe the benefits for which the community pursues toward this goal, the interested reader could follow more specific modern bibliography \cite{Wiersma2013,chaigne2014controlling,Frostig2017}. With our work, instead, we want to offer a novel way to approach the estimation of the transmission matrix via a random sampling statistical approach. The state-of-art for the calculation of the transmission matrix (from now on $\mathbb T$) is the method introduced by Popoff et al. \cite{Popoff2011}, which uses the Hadamard basis as input and reconstruct the $\mathbb T$ based on output observations. Although the protocol has proven excellent performance, one would prefer a less strict sampling approach, ideally not relying on any input bases. In fact, the method works only in the forward direction (input to output) and thus gives access only to the direct transmission matrix. Operatively, we need to invert such $\mathbb T$ before being able to control the turbid device. Matrix inversion is not a trivial process, especially in the presence of noise rendering the system very sensitive to perturbation, where it is often preferable other approaces \cite{penrose1956best}. The need for a procedure that is robust and efficient  also for random sampling datasets is the reason why we approach statistics.
 Breaking the asymmetric approach in \cite{Popoff2011}, further might give access to both the reconstruction of the direct $\mathbb T$ and its inverse $\mathbb T^{-1}$ with the same technique, opening up novel possibilities to calculate correlations. Our study fits in this context, inferring how we learn the transmission matrix by making use of statistical inference tools. 
 
 The model we propose in the following is directly derived from the linear propagation of light in arbitrary media \cite{Tyagi16} and turns out to be a spin-glass-like model with input/output (I/O) variables, rather then spins.
  This translation let us use a number of statistical tools to tackle the problem of the $\mathbb T$-parameters inference. In particular, we make use of the pseudolikelihood maximization approach \cite{Ravikumar10,Ekeberg13} exploiting a decimation strategy \cite{Decelle2014}, to estimate and select the most representative parameters of the I/O model considered. The usage of the pseudolikelihood renders the problem independent, linearly scalable (with the number of parameters) and trivially parallelizable on GPU hardware. Last but not least, our approach could be seen as a thermodynamical alternative to any linear regression problem, thus leaving it generically applicable to any problem involving the solution of linear systems.

\begin{figure}[h]
\centerline{\includegraphics[height=4.5cm]{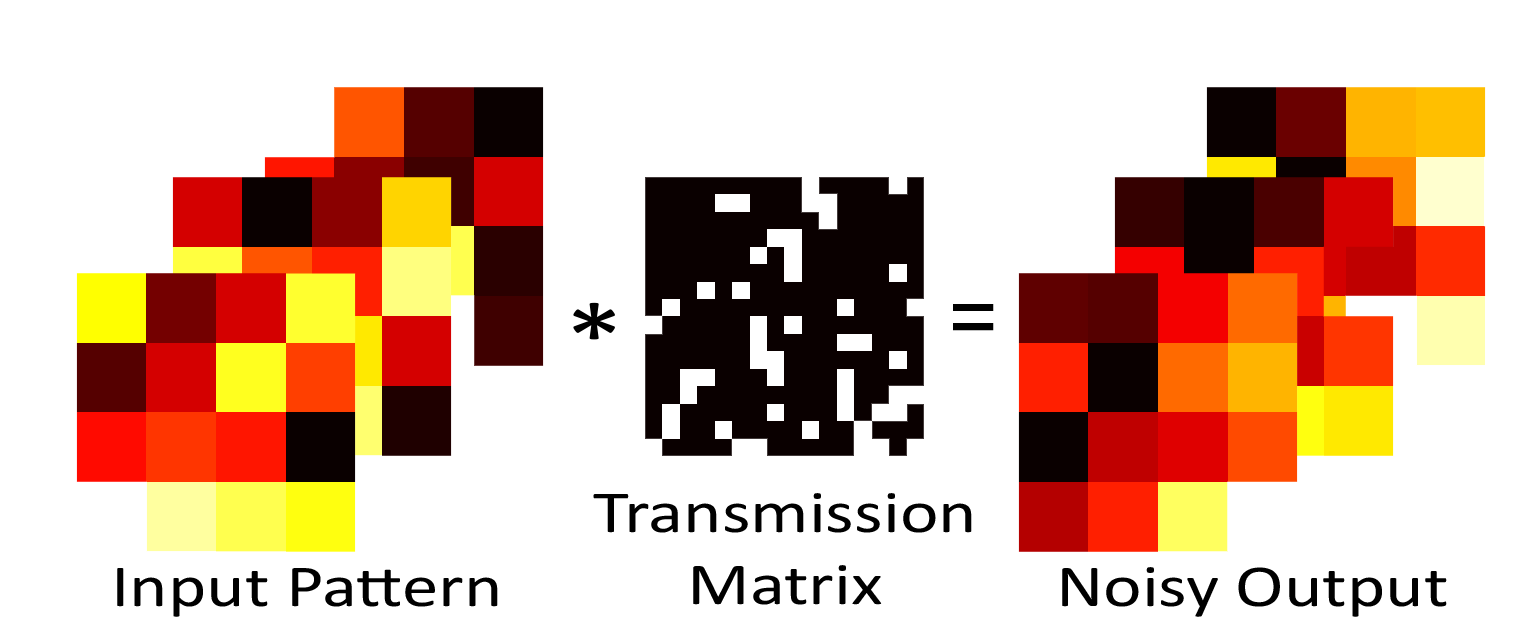}}
\caption{Schematics of the numerical experiment considered. We want to transmit an input signal trough a channel characterized by an unknown linear response. The response is fully encoded in the transmission matrix of the system, that gives access to the correct exploitation of the communication channel in both directions.}
\label{fig:Fig1_schematics}
\end{figure}

\section{Theoretical Framework}
\label{sec:mathfram}
Let us start considering a linear problem with an input-output transmission channel. To keep the analogy with disordered photonics we could see it as an I/O system in which a random medium (such as a multimode fiber) acts as a linear light scrambler: any input vector of intensity distribution $\bm{I}^{\rm in} = \{I_\eta^{\rm in} \}$, $\eta = 1, \ldots, N_I$, is connected to the output pattern $\bm{I}^{\rm out} = \{I_\gamma^{\rm out} \}$, $\gamma = 1, \ldots, N_O$,
 via an intensity transmission matrix $\mathbb T$ by the rule given by:
\begin{equation} \label{eq:transmission}
I^{\rm out}_\gamma = \sum_{\eta=1}^{N_I} T_{\gamma\eta}I_\eta^{\rm in} +  \epsilon_\gamma.
\end{equation}
Here, we assume for simplicity  the number of input  modes to be equal to the number of modes recorded at the output $N_O=N_I=N/2$, with $N=N_I+N_O$ and, thus, $ \gamma, \eta = 1,..., N/2$. We introduced a noise term   $\epsilon_\gamma$ in the last term of eq. \ref{eq:transmission}, to be able to include possible statistical perturbations in the measurement, due, e. g., to stochastic temperature changes or to bending or vibrations of the fiber. 
Such noise term is a random number of zero mean and finite mean square displacement $\sigma_\gamma$. 
In vector notation, eq. (\ref{eq:transmission}) can be compactly written as $\bm{I}^{\rm out} = \mathbb T \bm{I}^{\rm in} + \bm{\epsilon}$.
Initially, we will impose $\sigma_\gamma=0,\ \forall \gamma$.

We point out that eq. (\ref{eq:transmission}) describes intensity interactions. It would be exact in the case of imaging incoherent sources (such as fluorescence, phosphorescencem, black body radiation, incandescent lamps, etc.). Although similar equations were used for image reconstruction through fiber bundles \cite{kim2014toward} in interferometric configuration, a rigorous appproach would consider complex electromagnetic fields $\bm{E}^{\rm out} = \mathbb{T}_{E} \bm{E}^{\rm in} + \bm{\epsilon}$ \cite{Popoff11,Bianchi12}. In any case, though, we would eventually be bound to learn $\mathbb{T}$ from measures of their intensities $\bm I = \norm{\bm E}^2$, due to technological limitation, thus an intensity-only model is the only one  apt to infer the transmission parameter.

A further implementation considering complex amplitudes as entries \cite{Tyagi16}  would not be qualitatively more complicated than the procedure that we propose here. Our work can be straightforwardly generalized in presence of coherent sources and data available for input and output complex amplitudes (i. e., intensities {\em and} phases).
More complicated, in the case of intensity-only data, is an approach taking into account the phases and our lack of their knowledge: a  more strict approach, also much more computationally demanding, would be to average over the unknown phase degrees of freedom with their probability distribution \cite{Ancora20}. As we will see, under the assumptions of validity of the present approach, this would be the equilibrium statistical mechanic Boltzmann-Gibbs distribution. 
In this work, however, we limit ourselves to the model Eq. \ref{eq:transmission} already able to yield very effective estimates at a rather low computational cost.
 Thus, in the present representation, $\mathbb T$ is a matrix connecting the intensities between the fiber's ends, it is not the exact electromagnetic transmission matrix of the fiber. In the article we refer to it as effective-$\mathbb T$, or simply transmission matrix.


Let us now build the inference model. Satisfying the system equations (\ref{eq:transmission}), at zero noise, means requiring that
\begin{equation} \label{eq:delta}
\int \prod_{\gamma=1}^{N/2} dI_\gamma^{\rm out} \delta \left( I^{\rm out}_\gamma - \sum_{\eta=1}^{N/2} T_{\gamma \eta}I_\eta^{\rm in} \right) = 1
\end{equation}
for any given entry $\bm I^{\rm in}$.
 The integral  (\ref{eq:delta}) can be rewritten expressing each $\delta$-function with a Gaussian in the limit of its standard deviation 
 $\sigma_\gamma$ going to zero:
\begin{equation} \label{eq:delta_lim}
\lim_{\bm \sigma\rightarrow \bm{0}} 
\int\prod_{\gamma=1}^{N/2}    \frac{dI_\gamma^{\rm out}}{\sqrt{2\pi\sigma_\gamma^2}} \exp\left\{ -\frac{1}{2\sigma_\gamma^2}  {\left(I^{\rm out}_\gamma - \sum_{\eta=1}^{N/2} T_{\gamma\eta}I_\eta^{\rm in}\right)^2}\right\}= 1
\end{equation}
Keeping $\sigma_\gamma$'s finite amounts to introduce Gaussian noise, recovering
 the noise statistics introduced in eq. (\ref{eq:transmission}) for a perturbative term following a Gaussian distribution. 
 For the sake of generality we consider different noise for each output channel $\gamma$.
 If $\sigma_\gamma=\sigma$, $\forall \gamma$
  The exponent in eq. (\ref{eq:delta_lim}) can be expressed as a bilinear cost function:
 \begin{align}
\mathcal{H} &= \sum_{\gamma=1}^{N/2}  \left(I^{\rm out}_\gamma - \sum_{\eta=1}^{N/2} T_{\gamma\eta}I_\eta^{\rm in}\right)^2 = \nonumber \\
&= \sum_{\gamma=1}^{N/2} \left(I_\gamma^{\rm out}\right)^2
 - 2\sum_{\gamma, \eta}^{1,N/2} I_\gamma^{\rm out} T_{\gamma\eta} I_\eta^{\rm in} 
 + \sum_{\gamma, \eta,\eta'}^{1,N/2} T_{\gamma\eta}T_{\gamma\eta'}I_\eta^{\rm in}I_{\eta'}^{\rm in} \nonumber \\
&\equiv - \sum_{i,j} I_i J_{ij} I_j  = -\bm{I}^T \mathbb{J} \bm I
 \label{eq:hamiltonian}
\end{align}
where we define the concatenated input-output intensity vector $\bm I =\{ I_{\gamma}^{\rm in},I_{\eta}^{out} \}$ and new indices are defined such that $i,j = \{ \gamma, \alpha + N/2 \} = \{1, .., N\}$. 
The interaction matrix $\mathbb{J} = \{J_{ij}\}$ has the form of a four-block
tensor containing the transmission matrix $\mathbb{T}$ in the out-of-diagonal blocks:
\begin{equation} \label{eq:generalizedJ}
\mathbb{J}=
\begin{pmatrix}
    -\mathbb{U}	& \mathbb{T}^\dagger \\
    \mathbb{T}	& -\mathbb{I} \\
\end{pmatrix}
\end{equation}
Here, we can recognize also the input self-coupling matrix $\mathbb{U}=\mathbb{T}^\dagger \mathbb{T}$, and the identity matrix $\mathbb{I}$. 

The latter form of eq. (\ref{eq:hamiltonian}) is similar to the spin-glass Hamiltonian.
In the statistical mechanics formulation  $\mathcal{H}$ is the Hamiltonian
 of the system of which the interaction matrix $\mathbb{J}$ -structured as in eq. \ref{eq:generalizedJ}- is the subject of our inference procedure. 
Writing $\beta = 1/2\sigma^2$, the stationary probability to have an input-output configuration $\bm I$ with a certain energy given the set of interaction $\{J_{ij}\}$ is the Boltzmann distribution:
\begin{equation} \label{eq:probability}
P(\bm I | \mathbb{J}) = \frac{1}{\mathcal{Z}(\mathbb{J},\bm I)} e^{-\beta \sum^{1,N}_{ij} I_i J_{ij} I_j}.
\end{equation}
In this context, the normalization $Z$ is the canonical partition function that normalizes the probability distribution:
\begin{align} \label{eq:partition_function}
\mathcal{Z}(\mathbb{J},\bm I) = \left( \frac{1}{2\pi\sigma^2} \right)^{N/2} \int  \prod_{j=1}^N dI_j~  e^{-\beta \mathcal{H}(\mathbb{J},\bm I)}.
\end{align}
Having defined the probability, it is straightforward to write the log-likelihood $\mathcal{L}$ of the system. The function $\mathcal{L}$ describes how the set of interaction parameters $\{J_{ij}\}$ are likely to describe the observed intensity configuration $\bm I$:
\begin{equation}
\mathcal{L}(\bm I | \mathbb{J}) =  \ln \big[ P(\bm I | \mathbb{J}) \big]
\end{equation}

Applied to our case, the Maximum Likelihood Estimation (MLE) principle states that the interaction matrix $\mathbb J_{\rm inf}$ that maximize the function $\mathcal{L}$ is the one that most likely describe the system interactions:
\begin{equation}
\mathbb J_{\rm inf} = \arg \max \mathcal{L}(\bm I | \mathbb{J})
\end{equation}

We rely on the MLE principle to infer the transmission matrix, introducing a common approximation in the following, due to the fact that $\mathcal{L}$ is very difficult to maximize as the number of degrees of freedom increases.

\subsection{\label{ssec:pslIO}Pseudolikelihood of the I/O model}
Although there are a few exceptions, maximize $\mathcal{L}$ is challenging. Its computational  complexity grows exponentially with the number of parameters, mostly due to the partition function. To have access to analytical solutions linearly scalable with the system size, it is legitimate to introduce the log-pseudolikelihood function $\mathcal{L}_i$ \cite{barber2012bayesian} of the $i$-th element. 
We follow a common approach tested in spin-glass inference \cite{Tyagi2016,Marruzzo2017a}, looking at the probability of the i-th element conditional to the realization of the others $\backslash i$:
\begin{equation}
P(I_{i}|\bm I_{\backslash i}) = \frac{P(\{I_i, \bm I_{\backslash i}\})}{P(\bm I_{\backslash i})}.
\end{equation}
where $\bm I_{\backslash i}$ is the set of all intensities but $i$.
Before  proceeding with the calculation of the probability $P(I_{i}|\bm I_{\backslash i})$, it is useful to explicitly separate the Hamiltonian contributions involving the variable $i$, $\mathcal{H}_i$.

Starting from Eq. (\ref{eq:delta_lim}), for a given output pixel variable $\gamma=1,\ldots,N/"$,
we can write its  contribution to the (temperature rescaled) Hamiltonian as:
\begin{equation}
\label{eq:single_out_h}
H_\gamma \equiv  \beta_\gamma \mathcal{H}_\gamma(I_\gamma^{\rm out}| I^{\rm out}_{\backslash \gamma}, \bm I^{\rm in}) = 
\beta_\gamma\left(I^{\rm out}_\gamma\right)^2 -
2\beta_\gamma I^{\rm out}_\gamma\sum_{\eta=1}^{N/2}T_{\gamma\eta} I^{\rm in}_\eta + \beta_\gamma\sum_{\eta\eta'}^{1,N/2} T^\dagger_{\eta\gamma}T_{\gamma\eta'} I^{\rm in}_\eta I^{\rm in}_{\eta'}  \ ,
\end{equation}
where we have explicitly considered a channel dependent ``inverse" noise 
\begin{equation}
\label{def:betagamma}\beta_\gamma \equiv \frac{1}{2\sigma_\gamma^2} \ .
\end{equation}
We also 
define the related partial canonical partition function for the single variable Hamiltonian, given all the other variable values as quenched,  as:
\begin{eqnarray} \label{eq:partitionfunction}
\mathcal{Z}_\gamma &=& \int d I^{\rm out}_\gamma 
e^{-  H_\gamma(I_\gamma^{\rm out}| I^{\rm out}_{\backslash \gamma}, \bm I^{\rm in}) }
\end{eqnarray}

Analogously, the (noise-rescaled)  input variable  Hamiltonian contributions relevant to the pseudolikelihood maximization will have the form:
\begin{eqnarray}
\label{eq:single_in_h}
H_\eta(I_\eta^{\rm in}| I^{\rm in}_{\backslash \eta}, \bm I^{\rm out}) & \equiv&  -2 I^{\rm in}_\eta\sum_{\gamma=1}^{N/2}  I^{\rm out}_\gamma  \beta_\gamma T_{\gamma\eta} 
+ 2 I^{\rm in}_\eta \sum_{\xi< \eta} V_{\eta\xi} I^{\rm in}_{\xi} + \left( I^{\rm in}_\eta\right)^2 V_{\eta\eta}
\\
V_{\eta\xi}&\equiv&  \sum_{\gamma} \beta_\gamma T^\dagger_{\eta\gamma}T_{\gamma\xi}
\label{def:V}
\\
\mathcal{Z}_\eta &=&  \int d I^{\rm in}_\eta e^{-H_\eta(I_\eta^{\rm in}| I^{\rm in}_{\backslash \eta}, \bm I^{\rm out})}
\end{eqnarray}

We can write the pseudo-likelihood distribution of the generic $i$-th variable (input or output) given the 
 parameters $\mathbb{V}$, $\mathbb{T}$ and $\{\beta_\gamma\}$  and the intensities $\bm I_{\backslash i}$ as:
\begin{equation}
P\left(I_{i}|\bm I_{\backslash i}\right) = \frac{e^{- H_i}}{\mathcal{Z}_i} \quad , \quad i=1, \ldots, N
\end{equation}
and define the related log-pseudolikelihood per each element $i$ as:
\begin{equation} \label{eq:psl}
\mathcal{L}_i = \ln P\left(I_{i}\big| \bm I_{\backslash i}\right)  =- H_i  - \ln \mathcal{Z}_i
\end{equation}

In general the maximum likelihood estimation of each single variable pseudo-likelihood function $\mathcal{L}_i$ is carried out searching the parameter's space.  In this way the same parameter, say $V_{\eta \xi}$, will be inferred twice, from MLE of $\mathcal{L}_\eta$ and  from  MLE of $\mathcal{L}_\xi$. Since its values are, usually, different, its average value is taken as estimate. 
Here, we will exploit our knowledge of the symmetry properties of the system in maximizing the sum of all pseudolikelihoods:
\begin{align}
\mathcal{L} = \sum_{i=1}^{N} \mathcal{L}_i
\label{eq:totalPSL}
\end{align}
that we refer to as the total log-pseudolikelihood function.
Without loosing specificity, we drop the "log-" prefix and we refer directly to pseudolikelihood ($\mathcal{L}$) functions.

To put together Eq. (\ref{eq:single_out_h}) and Eq. (\ref{eq:single_in_h}) in Eq. (\ref{eq:totalPSL}) it can be convenient to
introduce  $N/2\times N/2$ diagonal matrix $\bm\beta$, whose elements are the ``inverse temperatures" $\{\beta_\gamma\}$,
and  
to give the following generic 
expression in terms of $\bm I = \{\bm I^{(\rm in)}, \bm I^{(\rm out)}\}$:
\begin{eqnarray} 
\label{eq:sep_hamiltonian}
H_i &=& I_i {\sum_{j \neq i}^{1,N} M_{ij} I_j  + I_{i}^2 M_{ii}}   = I_i B_i - I_i^2 A_i , \ \quad i = 1, \ldots, N
\end{eqnarray}
with 
\begin{eqnarray} 
\label{def:M}
\mathbb{M}&\equiv& \begin{pmatrix}
   \Ddot{15}.(5pt,10pt,13pt,-5.3pt) \phantom{-\mathbb{V}\mathbb{I}}	& & -2\mathbb{V} & \!{|}  \,  \phantom{   -\mathbb{VI}}	 & & \,  \,  \,   \,   \,  \phantom{   -\mathbb{V}\mathbb{I}}
    \\ & -\mathbb{\hat V} & &\! {|} \,  \phantom{   -\mathbb{V}\mathbb{I}}	& 2\mathbb{T}^\dagger \bm\beta  &
        \\ -2\mathbb{V}  & &\! \phantom{ -\mathbb{V}\mathbb{I}} & \!{|} \,  \phantom{   -\mathbb{V}\mathbb{I}}	& &
        \\ \hline
     & & &\!| \,  \phantom{   -\mathbb{V}\mathbb{I}}	 & &
    \\ &  2 \bm\beta  \mathbb{T} & &\!| \,  \phantom{   -\mathbb{V}\mathbb{I}}	&\, -\bm\beta  \,&
    \\
   	&   & &\!| \,  \phantom{   -\mathbb{V}\mathbb{I}}	& &   \\
\end{pmatrix}
\\
\label{eq:parameters}
\quad A_i &\equiv& -M_{ii} \quad \ , \qquad B_i \equiv \sum_{j \neq i}^{1,N} M_{ij}I_j \ .
\end{eqnarray}
where by $\mathbb{\hat V}$ we mean the diagonal part of $\mathbb{V}$. The partition function is, then, written as:
\begin{eqnarray} 
\mathcal{Z}_i &=& \int dI_i~ e^{-H_i}
\label{eq:Zi}
\end{eqnarray}
By definition the diagonal part of $\mathbb{M}$ is always negative, cf. Eqs. (\ref{def:V}) and (\ref{def:M}): $A_i =\sum_{j=1}^{N/2} \beta_j T_{ji}^2 > 0$, for $i=1,\ldots,N/2$ and 
$A_i= \beta_{i-N/2}>0$, for $i=N/2+1, \ldots, N$, so that  the partition function converges without worrying about the intensity saturation cut-off, that is, the dominion of $I_i$ in integral Eq. (\ref{eq:Zi}).

We stress that, since $\mathbb{M}$ is symmetric, only the elements $M_{i\leq j}$ will be inferred, since there are at most $N(N+1)/2$ independent terms.
In case of sparse transmission matrices the MLE of the $\mathcal{L}$ is, then, followed by a decimation procedure \cite{Decelle2014}, in which small couplings are recoursively set to zero until $\mathcal{L}$ is stable.
 
\subsubsection{\label{ssec:partitionfunction}Which partition function?}
It is obvious that  $\mathcal{L}_i$  depends upon the choice of the integration range in the partition function. This choice is bound to the values each $I_i$ can take. First of all, it is possible to analytically solve the indefinite integral \ref{eq:Zi}, obtaining:
\begin{align}
\mathcal{Z}_i(I_i) = \sqrt{\frac{\pi}{4A_i}} e^{\frac{B_i^2}{4A_i}} \erf {\left( \frac{2A_i I_i -B_i}{\sqrt{4A_i}} \right)}
\equiv \sqrt{\frac{\pi}{4A_i}} e^{\frac{B_i^2}{4A_i}} \mathcal{F}(I_i).
\end{align}
\begin{figure}[h!]
\centerline{\includegraphics[height=10cm]{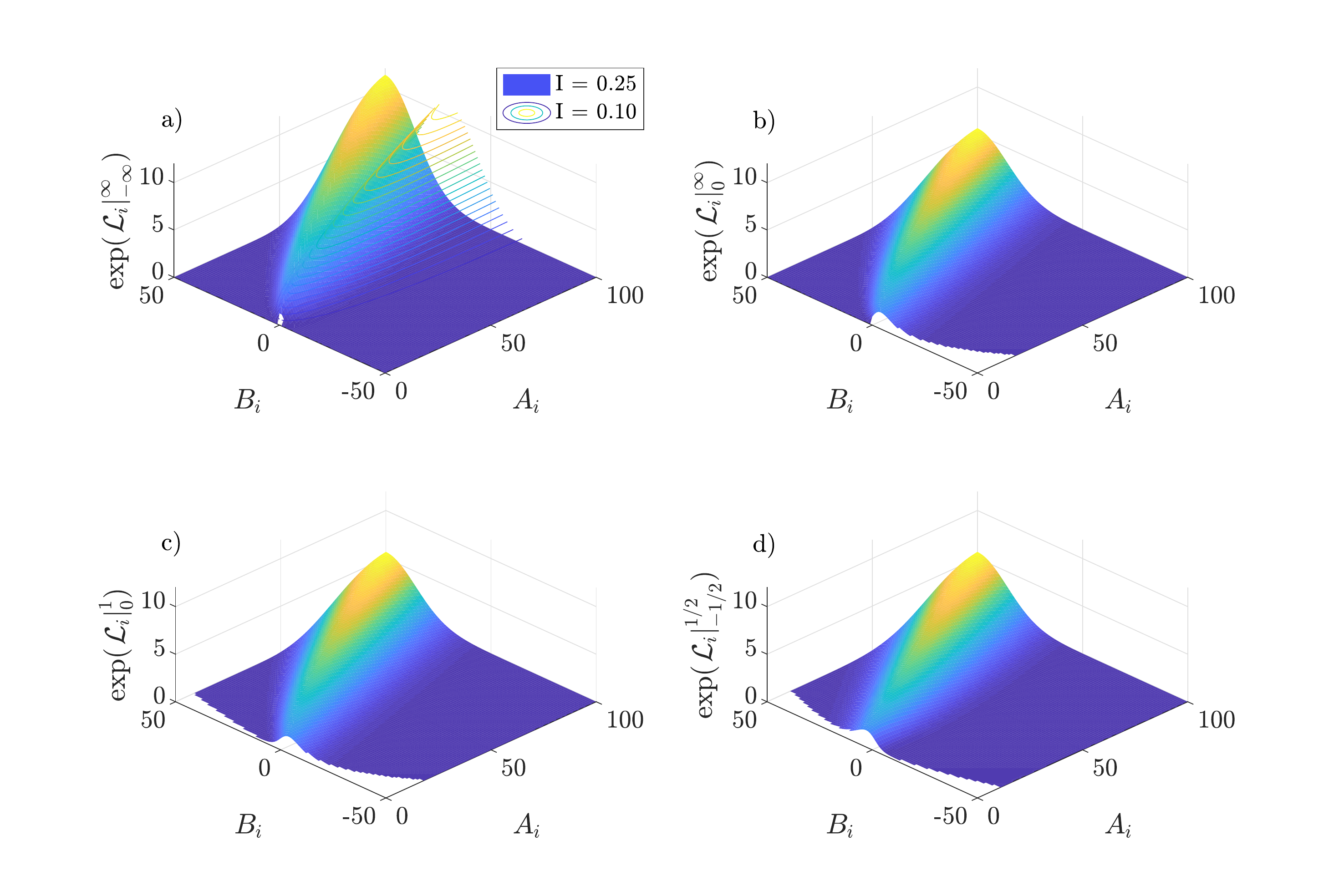}}
\caption{Pseudolikelihood for the $i$-th intensity elements given all the others. The function is plotted against the parameters $A_i$ and $B_i$, using   $I_i = 0.25$ as a reference value for the intensity. In the first plot we draw the contours lines for a different case $I=0.10$. Although the plots differ in terms of absolute values, it is possible to notice how the maximum follow the same direction in all the cases.}
\label{fig:probabilityfunctions}
\end{figure}
The only term dependent on $I_i$ is the argument of  $\mathcal{F}$, which changes upon different choice of the integral limits $[I_{min}, I_{max}]$ as, for instance:
\begin{equation}
\mathcal{F}_i = 
\begin{cases}
  \ 2   & \text{for }  \left. \mathcal{F}_i \right| ^\infty_{-\infty} \\
  \ 1 + \erf \left( \frac{B_i}{\sqrt{4A_i}} \right) & \text{for } \left. \mathcal{F}_i\right| ^\infty_0 \\
  \ \erf \left( \frac{2A_i - B_i}{\sqrt{4A_i}} \right) + \erf \left( \frac{B_i}{\sqrt{4A_i}} \right) & \text{for } \left. \mathcal{F}_i\right| ^1_0 \\
  \ \erf \left( \frac{ 2A_i -B_i}{\sqrt{4A_i}} \right) + \erf \left( \frac{ 2 A_i + B_i}{\sqrt{4A_i}} \right) & \text{for } \left. \mathcal{F}_i\right| ^{1}_{-1} \\
\end{cases}
\label{def:F}
\end{equation}
We stress that the intensities of the pixels are defined as positive quantities. As we will see in the following, though, letting them run along negative values, as well, does not alter the inference quality and allows for a simplified computation.
The correct selection of the integral limits should take into account all the possible intensities allowed in the system. In this framework, the $\mathcal{L}_i$ can be explicitly written as:
\begin{align} \label{eq:pseudolikelihood_i}
\mathcal{L}_i = I_i B_i - I_i^2A_i - \frac{1}{2} \ln \left( \frac{\pi}{4A_i} \right) - \frac{B_i^2}{4A_i} -\ln \left. \mathcal{F}_i\right| ^{I_{max}}_{I_{min}}.
\end{align}
To have a graphical view on the various functions introduced in eq. (\ref{eq:pseudolikelihood_i}), we plot the generic pseudolikelihood as a function of the parameters $\exp(\mathcal{L}_i)=P(A_i,B_i|I_i)$ at given $I_i$, for different choices of the integration extremes in Fig. \ref{fig:probabilityfunctions}. We can qualitatively appreciate how the maxima of the functions follow the same direction along the plane $(A_i,B_i)$. As $I_i$ varies the maximum of $P(A_i,B_i|I_i)$ varies \footnote{We do not have a unique maximum, since it will be given by the choice of the $\{J_{ij}\}$ rather than the two parameters of eq. (\ref{eq:parameters})}.
We consider four cases though  the simplest choice of $\left.\mathcal{F}_i \right| ^\infty_{-\infty} = 2$ turns out to be  operatively the most efficient for all the intensities tested. 


\subsection{\label{ssec:derivatives}Pseudolikelihood Derivatives}
The formulation of the pseudolikelihood (\ref{eq:pseudolikelihood_i}) defines a concave function that admits (under low enough noise) a unique maximum.  To ease the maximization process is, thus, useful to calculate its first derivatives. Let us first compute the derivatives of the parameters $A'_i$ and $B'_i$, defined in eq. (\ref{eq:parameters}), to simplify the notation:
\begin{align}
A_i' = \frac{\partial A_i}{\partial M_{nm}} = - \delta_{in} \quad \quad B_i' = \frac{\partial B_i}{\partial M_{nm}} = \delta_{in} \delta_{jm} I_j.
\end{align}

It is now possible to calculate the derivatives of $\mathcal{L}_i$ with respect to the parameters of the interaction matrix $\mathbb{M}$:
\begin{align}
\frac{\partial \mathcal{L}_i}{\partial M_{nm}} &= I_i B_i' - \frac{1}{\left[\mathcal{F}_i\right]} \frac{\partial \mathcal{F}_i}{\partial M_{nm}}, \\
\frac{\partial \mathcal{L}_i}{\partial M_{nn}} &= -I_i^2 A_i' + \frac{A_i'}{2A_i} - \frac{1}{\left[\mathcal{F}_i\right]} \frac{\partial \mathcal{F}_i}{\partial M_{nn}}.
\end{align}

Of course, the derivatives of $\mathcal{F}_i$ depend upon the choice of the integration limits thus, for the cases analyzed in \ref{ssec:partitionfunction}, we have that the out of diagonal terms of $\mathbb{M}$ can be expressed as:
\begin{equation}
\frac{\partial \mathcal{F}_i}{\partial M_{nm}} = B_i' \left( \frac{e^{-\frac{B_i^2}{4A_i}}}{\sqrt{\pi A_i}} \right)
\begin{cases}
  \ 0   & \text{for }  \left. \mathcal{F}_i \right| ^\infty_{-\infty} \\
  \ 1 & \text{for } \left. \mathcal{F}_i\right| ^\infty_0 \\
  \ 1-e^{-A_i+B_i} & \text{for } \left. \mathcal{F}_i\right| ^1_0 \\
  \ e^{-A_i-B_i}-e^{-A_i+B_i} & \text{for } \left. \mathcal{F}_i\right| ^1_{-1} \\
\end{cases}, \ \forall n \neq m
\end{equation}
and the diagonal part  gives:
\begin{equation}
\frac{\partial \mathcal{F}_i}{\partial M_{nn}} = - \frac{A_i'}{2A_i} \left(\frac{e^{-\frac{B_i^2}{4A_i}}}{\sqrt{\pi A_i}} \right)
\begin{cases}
  \ 0   & \text{for }  \left. \mathcal{F}_i \right| ^\infty_{-\infty} \\
  \ B_i & \text{for } \left. \mathcal{F}_i\right| ^\infty_0 \\
  \ B_i - e^{-A_i+B_i} \left(B_i + {2A_i} \right) & \text{for } \left. \mathcal{F}_i\right| ^1_0 \\
  \ e^{-A_i-B_i} \left(B_i - {2A_i} \right) - e^{-A_i+B_i} \left(B_i + {2A_i} \right)  & \text{for } \left. \mathcal{F}_i\right| ^1_{-1} \\
\end{cases}.
\end{equation}

These derivatives are a sufficient requirement to use quasi-Newtonian methods for finding the maximum of a function, without explicitly calculating the Hessian. However, second derivatives can be explicitly calculated, even if the algorithm should pose dedicated care to avoid out-of-memory issues.


\subsection{Measurement Sampling}
\label{sec:MeaSam}
Having set a statistical framework, we have to consider the quantities introduced above as averaged over $M$ measurements:
\begin{equation}
\mathcal{L} = \frac{1}{M} \sum_{i=1}^N \sum_{\mu=1}^M \mathcal{L}_{i, \mu}
\end{equation}
This depends on the $N\times N$ $M_{ij}$ coupling parameters. However, for the symmetry properties of Eq. (\ref{eq:sep_hamiltonian}),
only $N(N+1)/2$ of them are independent. We maximize $\mathcal{L} $ only with respect to those, forming the lower triangle $M_{ij}$, $i\leq j$, of matrix 
 (\ref{def:M}), including the diagonal.
The $N(N+1)/2$ derivatives are, then,
\begin{align}
\frac{\partial  \mathcal{L} }{\partial M_{nm}} &= \frac{1}{M} \sum_{i=1}^N \sum_{\mu=1}^M \frac{\partial \mathcal{L}_{i,\mu}}{\partial M_{nm}} 
\end{align}
where $M$ is the number of coupled I/O observations of a system with a fixed $\mathbb{T}$. The sampling rate of a given system will impact on the inference procedure: a larger statistical sampling will correspond to a more accurate learning process, i. e., to a better estimate of the model parameters. Thus, we define the sampling rate $\xi(M,K)$ as:
\begin{equation}
\label{def:samp_rate}
\xi(M,K) = \frac{M}{K}
\end{equation}
where $\xi$ depends on the number of parameters $K$ to be inferred in the model and on the parameters. Three situations can occur in this case, depending on the sampling rate achieved:
\begin{list}{•}{}
\item $\xi<1$, undersampling regime: the number of measurements is lower than the number of parameters, worst inference scenario.
\item $\xi \gtrsim 1$, sufficient sampling: $M$ approximately matches $K$, thus the sampling rate is appropriate for a stable learning procedure.
\item $\xi\gg 1$, oversampling regime: best case scenario, in which the number of measurements is high enough to reasonably rule out pre-asymptotic finite measurement effects in the convergence of the pseudolikelihood to the actual likelihood function.
\end{list}

Ideally, we would like to be able to correctly infer the parameters in all the possible sampling rates, in particular in the undersampled regime.  In the following, we will try to analyze the behaviour of our learning framework under different measurement scenarios.

\section{Numerical Implementation}
\label{sec:numimpl}
The algorithm is written in MATLAB, making use of the unconstrained minimization routine minFunc \cite{schmidt2005minfunc} enjoying the L-BFGS algorithm, which uses a low-rank Hessian approximation. This lets the system accept a higher number of variables, keeping the lowest memory usage. Of course we are not bound to use L-BSGF algorithms for the optimization, but we found it a good compromise between accuracy, speed and memory requirement. We also have tested the use of constrained optimization since we have full control on the parameter range in the system. However, we decided to report unconstrained results to be more general and to assess the robustness of the method.

\subsection{Dataset creation and its statistics}
\label{ssec:datasetstatistics}
To test the inference model, we created numerically a set of intensity measurements, to simulate a laboratory experiment. First of all, any real camera measurement returns a positive value for the intensity, limited in magnitude by the camera  sensitivity: values above the camera maximal sensitivity $I_{\rm max}$ will saturate and below the minimal sensitivity $I_{\rm min}$ will not be recorded. For the sake of generality, we rescale these values and we assume that our simulated camera works in the range $I_i \in \left[0, 1 \right]$ for both the input and the output. For the input, we choose to generate $M$ squared patterns of pixels having size $w$. Each of these contains $N_I=N_O=N/2=w^2$ random values normally distributed, accordingly to a truncated gaussian probability distribution with $\sigma^{\rm in}=0.1, \ \forall \gamma$ and $\mu=0.5$. We point out that $\sigma^{\rm in}$ is not related to the noise, but defines the distribution of intensity values sampled in the input. These patterns are transported into the corresponding outputs via the $\mathbb{T}$ matrix, that must preserve the positiveness and the intensity range. 
\begin{figure}[b!]
\centerline{\includegraphics[height=7cm]{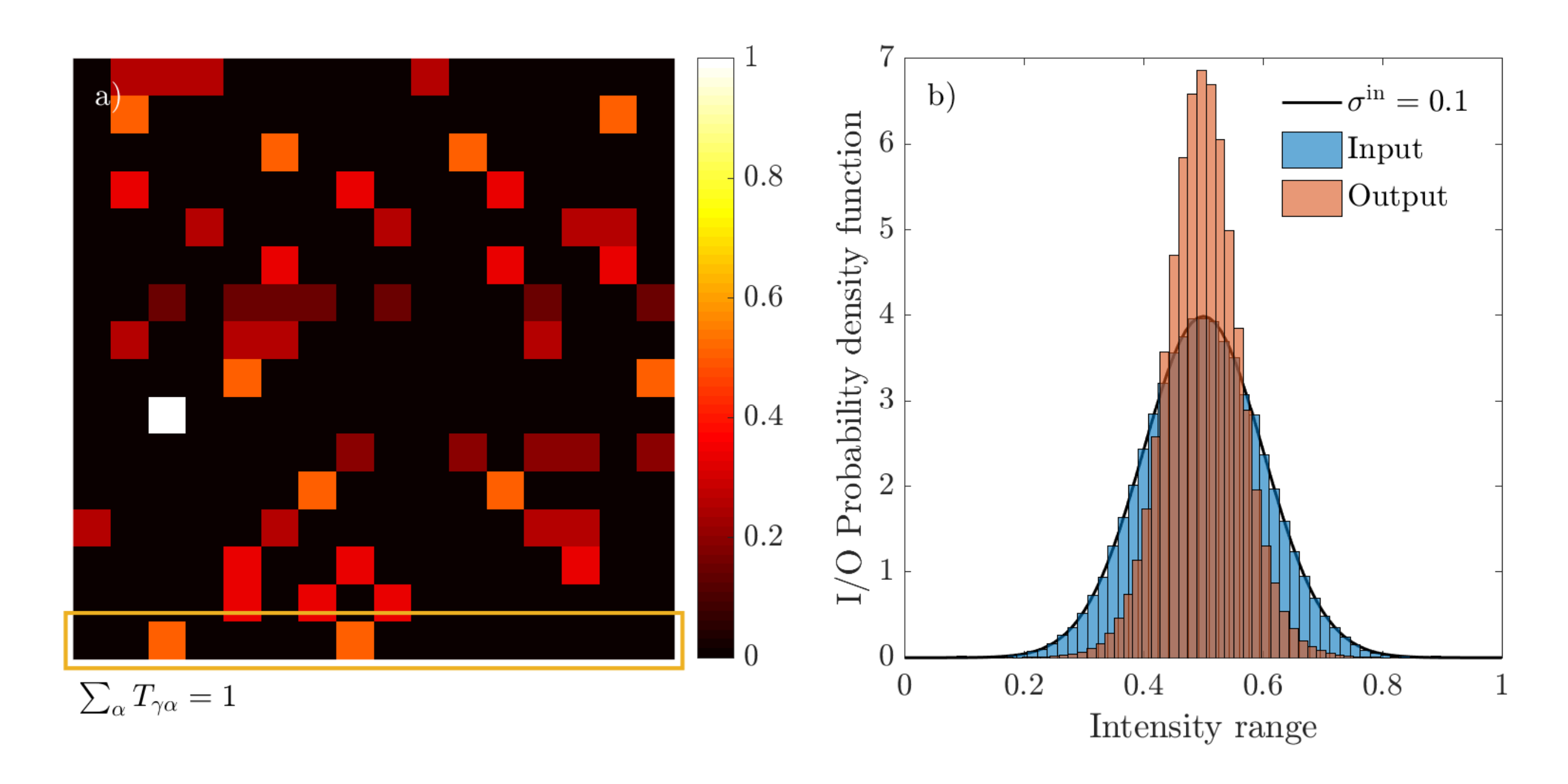}}
\caption{On the left, an example of a transmission matrix for a 4x4 input-output system. On the right the probability distribution chosen for all the pixel in the input (blue histogram) and the corresponding distribution after passing through the transmission channel (orange histogram). Both have gaussian statistics with mean value centered at $0.5$.}
\label{fig:Fig2_IOstatistics}
\end{figure}

In order to build our $\mathbb{T}$ we first decide that a fraction $s$ of all possible entries will be active (non-zero), i. e.,  $s = S_{\rm active} / w^4$. Given the sparsity, we randomly activate $S_{\rm active}$ couplings in a temporary (binary) $\mathbb{T}'$-matrix, setting their value to $1$. We want also the output to be in the same $[0, 1]$ range, thus, we need to normalize the $\mathbb{T}'$ row wise, so that the matrix eventually be a stochastic matrix:
\begin{eqnarray}
\nonumber
T_{\gamma \alpha} =\frac{T'_{\gamma \alpha}}{\sum_\alpha T'_{\gamma \alpha}}
\label{T_stoc} \quad , \quad 
\sum_{\alpha=1}^{N_O} T_{\gamma \alpha}=1
\end{eqnarray}
 With this choice, in every experiment performed we are guaranteed to have also the output in the same intensity range of the input, as it is the case for CCD camera acquisitions. In general, we found that the histogram of the intensities in the output followed a narrower Gaussian distribution, as it can be seen in Fig. \ref{fig:Fig2_IOstatistics}. Depending on which integral 
 (\ref{def:F}) is chosen for the inference,   when the region of integration is taken symmetric around zero (explicitly, when we use  $\left. \mathcal{F}_i \right| ^\infty_{-\infty}$ or  $\left. \mathcal{F}_i \right| ^{1/2}_{-1/2}$), it is convenient to translate the intensities in each channel by their mean $I_i - \mu_i$,
  The  reason for this will be discussed in section \ref{ssec:PLmin} and in the Results.
  
\begin{figure}[b!]
\centerline{\includegraphics[height=7cm]{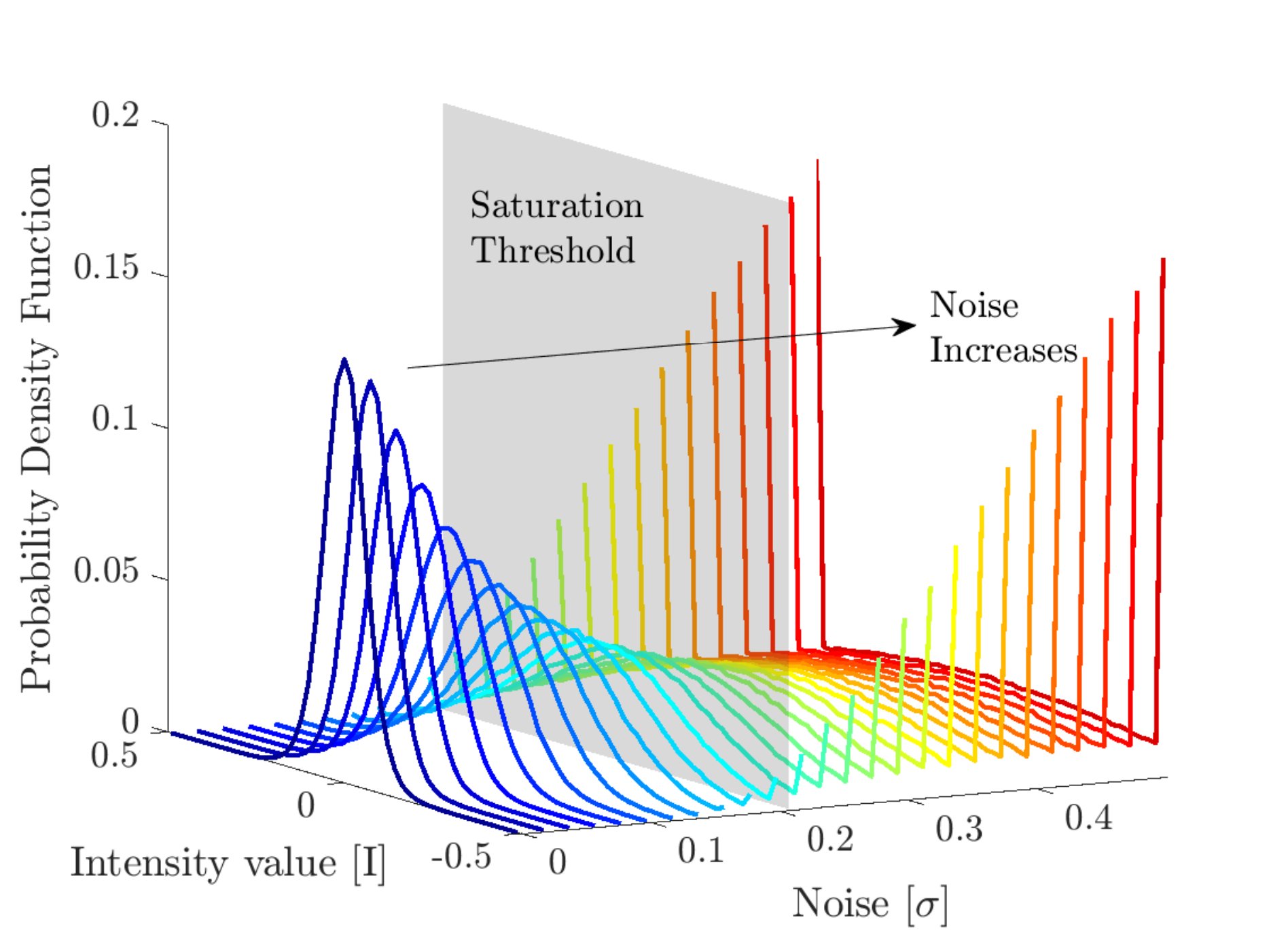}}
\caption{Statistics of the output intensities after perturbing the channel with noise at increasing variance. The jet colorbar on the right is syncronized with the noise variance. The effect is that the original distribution starts to widen and saturation effects appear at around $\sigma = 0.2$. The two wings represent an increased saturation effect due to very strong noise.}
\label{fig:Fig4_outputpdf_vs_noise}
\end{figure}

Accordingly to (\ref{sec:mathfram}), we will study the inference method under noisy measurements. We add a perturbative noise term to the output, leaving the input unaltered. The noise is a Gaussian random number with zero mean and variance $\sigma$. To probe different noise regimes we 
repeated our analysis changing $\sigma \in [0, 0.5]$, in steps of $0.02$, including possible strong saturation effects of the data.
Per each $\sigma$, we will run an optimization and consequently a full decimation of the parameters, studying the best model selection via the usage of several information criteria. The effect introduced by the noise perturbation is represented in Fig.\ref{fig:Fig4_outputpdf_vs_noise}; where it is clear that when the noise grows up to $20\%$, the channel is so perturbed that saturation effects become relevant.

\subsection{$\mathcal{L}$-maximization with all parameters}
We decided to start the optimization process using weak prior information. Recalling the tensor form of the $\mathbb{M}$, we initialize each block with the following criteria:
\begin{enumerate}
\item $\mathbb{I}$ block, we start with a diagonal unity matrix; 
\item $\mathbb{T}$ blocks, we set all the parameters to $1/w^2$;
\item $\mathbb{V}$ block, we calculate it as $\mathbb{T}^\dagger \bm{\beta}\ \mathbb{T}$ using the previously initialized matrix, with $\beta_\gamma=\beta=1/2/\sigma$, $\forall \gamma$.
\end{enumerate}

We call the initial coupling matrix as $\mathbb{M}^{(0)}$ and we run the minimization routine, obtaining the estimated couplings $\{M_{ij}\}$. We point out, that any initialization scheme should lead to the same optimal convergence for the parameters, but with the described choice we avoid potential wrong starting directions in the optimization process, leading to no or slow covergence. In particular, for the $\mathbb{I}$ region we immediately decimate the out-of-diagonal elements, that never take part in the  optimization. The resulting optimized interaction matrix is referred to as $\mathbb{M}_{\rm inf}$.

\subsection{Decimation strategy}
\label{ssec:decimationstrategy}
After the initial $\mathcal{L}$-maximization, we proceed decimating the region where the smallest couplings are inferred. Our assumption is that the transmission matrix has a certain degree of sparsity, which we want to recover during the learning procedure in order to avoid overfitting. We proceed iteratively, sorting the values of the inferred $\mathbb T$ and discarding a fraction, equal to $1/128$, of the smallest elements. Then, we repeat again the optimization starting with the previously inferred matrix.
 Ideally, the absolute value of the maximum $\mathcal{L}$ should follow a constant behaviour while decimating irrelevant couplings, while a decrease is expected when discarding relevant  parameters. The decimation strategy is directly applied on the $\mathbb T$-region via the definition of a binary parameter mask $\mathbb T^{\rm bin}$, which is $T^{\rm bin}_{ij} = 1$ when the coupling is active and $0$ otherwise. The decimation is transported into the $\mathbb V$-region by calculating another mask $\mathbb V^{\rm bin} =  (\mathbb T^{\rm bin})^\dagger  \bm \beta \ \mathbb T^{\rm bin}$. We restrict the decimation procedure only in the out-of-diagonal part of $\mathbb M$, thus leaving the diagonal parameters $\{M_{ii}\}$ acting as regularizers at advanced stages of decimation, in order to avoid divergence of the partition function. Per each decimation step, we call $K$ the total number of active parameters in $\mathbb{M}$.
 
Decimation was introduced in \cite{Decelle2014} to estimate couplings in an Ising model along with a stopping criterion, which allows us to estimate an appropriate number of parameters without proceeding further with the decimation. In this work, we decide to compare several information criteria (IC) that can be used to decide where to stop the decimation when a good $\mathbb M$ estimate is achieved. 

\begin{list}{•}{}
\item \textbf{Tilted-Pseudolikelihood Information Criterion (TIC):} Following the procedure introduced by Decelle and Ricci-Tersenghi in \cite{Decelle2014} we define the tilted-$\mathcal{L}$ information criterion (we call it TIC for nomenclature consistency with the others) as
\begin{equation}
{\rm TIC} = \mathcal{L} - k \mathcal{L}_{max} - (1-k) \mathcal{L}_{min}
\end{equation}
where $k = K / C_{total}$ is the fraction of the active and $(1-k)$ the fraction of the decimated couplings. $\mathcal{L}_{max}$ is the value achieved maximizing the fully connected model while $\mathcal{L}_{min}$ is the pseudolikelihood of the non-interacting model. 
 Ideally, we want to stop the decimation when $k^*=s$, the true fraction of non-zero elements in $\mathbb{T}$, as defined in Sec. \ref{ssec:datasetstatistics}.
 By definition, $k^{\rm TIC}$ is the value at which TIC is maximized, and $\mathbb{M}^{\rm TIC}$ is the optimized decimated interaction matrix.

\item \textbf{Akaike information Criterion (AIC):} The first criterion to estimate the optimal parameter number in a given model was introduced by Akaike's work \cite{akaike1974new}, as:
\begin{equation}
{\rm AIC} = 2K - 2 M \left(\mathcal{L} \right)
\end{equation}

The value $K^{{\rm AIC}}$ at which the AIC is minimized correspond to the model matrix $\mathbb{M}^{\rm AIC}$ that minimizes the information loss, thus, at the best parameter number representing the system.

\item \textbf{Corrected AIC (AICc):} When the sampling $M$ is small (though larger than $K$), there is a higher chance that  AIC will overfit, selecting models with a higher number of parameters. To take into account this, a modified version of the AIC, namely AICc, was introduced \cite{hurvich1989regression}:
\begin{equation}
{\rm AICc} = {\rm AIC} + \frac{2K \left(K+1\right)}{M-K-1}
\end{equation}
The above formulation takes into account the sampling number into a penalty factor, that looses importance when $M \rightarrow \infty$ converging to AIC value. The value $K^{\rm AICc}$ at which the AICc is minimized corresponds to a second-order estimate of the optimal model. $\mathbb{M}^{\rm AICc}$ is the corresponding decimated matrix.

\item \textbf{Bayesian Information Criterion (BIC):} Firstly introduced by Schwarz in \cite{schwarz1978estimating}, it can be written as a function of the estimable parameters $K$ and the number of sampling $M$ as:
\begin{equation}
{\rm BIC} = K \log (M) - 2 M  \mathcal{L} .
\end{equation}
The number of parameters $K^{\rm BIC}$ at which BIC is minimized is the best inference point accordingly to the Bayesian information criterion. Here, the interaction matrix selected is called $\mathbb{M}^{\rm BIC}$. This method resulted efficient in the activation method \cite{franz2019fast} for the inference of Ising couplings starting from empty matrices.
\end{list}

The information criteria introduced will help us to estimate the best model fitting the data, giving an indication on the sparsity and the corresponding number of parameters needed to correctly describe it. 

\subsubsection{$\mathcal{L}_{min}$ of the fully disconnected model. TBC}
\label{ssec:PLmin}
Accordingly with the decimation strategy proposed in \ref{ssec:decimationstrategy}, when the model is completely decimated only the values on the diagonal of the interaction matrix $\{M_{ii}\}$ are active. This implies that, for each $i$-th likelihood $\mathcal{L}_i$, the parameter $B_i = 0$ vanishes. In practice, we are left with:
\begin{equation}
\mathcal{L}_i \xrightarrow[\text{decimation}]{B_i \rightarrow 0} \ln{ \left( \frac{e^{M_{ii} I_i^2} }{\int d I_i e^{M_{ii} I_i^2}} \right) } = \ln{ \left( \frac{e^{-A_i I_i^2} }{\int d I_i e^{-A_i I_i^2}} \right) }
\end{equation}
The argument of the logarithm is a Gaussian distribution of variance $2/A_i$ with, cf. Eqs. (\ref{def:betagamma}),  (\ref{def:V}), 
\begin{equation}
- M_{ii} = A_i = \left\{ \begin{array}{cc}
 \frac{1}{2\sigma_{i-N/2}^2} & \quad \mbox{if } i =N/2+1, \ldots, N \ ,  \quad \mbox{output}
 \\
 \\
 \sum_{\gamma=1}^{N/2}\frac{1}{2\sigma^2_\gamma}  T^2_{\gamma i}& \mbox{if } i =1, \ldots, N/2 \ ,  \quad \mbox{input} \ .
\end{array}\right.
\end{equation}

When the model is fully decimated, maximizing each $\mathcal{L}_i$ corresponds to fit a Gaussian probability distribution on the distribution of the intensities per channel $i$. The main limitation is that the decimated model fits a Gaussian distribution fluctuating around a null value. For this reason, a theoretical calculation of the $\mathcal{L}_{min}$ is possible only if we shift $\bm I-\bm\mu$,  where $\bm\mu = \{\mu_i\}$ is the vector mean value calculated on the $M$ samplings. 
For the case described in section \ref{ssec:datasetstatistics} we have that $\mu_i = 0.5$ $\forall i$, thus it makes sense to shift all the intensities by this value. Let us recall, that in this case we need to use either $\left. \mathcal{L}_i \right| ^{+\infty}_{-\infty}$ or $\left. \mathcal{L}_i \right| ^{+1/2}_{-1/2}$, to accept negative values of the integration variables. This shift will not change the definition of $\mathbb{T}$. Now we can calculate the minimum value that $\mathcal{L}$ can reach as:

\begin{align}
\nonumber
 \mathcal{L}_{min} \Bigr| ^{{1/2}}_{{-1/2}} &= - \sum_{i=1}^N \left[ \frac{I_i^2}{2\sigma_i^2}+ \ln \left( \frac{\pi \sigma_i^2}{2} \right) + \ln \left. \mathcal{F}_i\right| ^{{+1/2}}_{{-1/2}} \right] \\
&= - \frac{1}{2} \sum_{i=1}^N \frac{I_i^2}{\sigma_i^2} - N \ln {\frac{\pi}{2}} - 2 \sum_{i=1}^N \ln{\sigma_i} - \sum_{i=1}^N \ln \left[ 2 \erf \left( \frac{1}{2\sqrt{2}\sigma_i} \right) \right]
\end{align}

or 
\begin{align} \label{eq:PL_min}
 \mathcal{L}_{min} \Bigr| ^{{\infty}}_{{-\infty}} &= - \frac{1}{2} \sum_{i=1}^N \frac{I_i^2}{\sigma_i^2} - \frac{N}{2} \ln {2 \pi} - \sum_{i=1}^N \ln{\sigma_i}.
\end{align}

In the scenario in which there are a variance characterizing the input $\sigma_I$ and one characterizing the output $\sigma_O$ equiparted between $N/2$ inputs and $N/2$ outputs, the formula above reduces to:
\begin{align}\label{eq:PL_min_simple}
\mathcal{L}_{min} = - \frac{1}{2} \left( \frac{1}{\sigma_I^2}\sum_{i=1}^{N/2} I_i^2 + \frac{1}{\sigma_O^2}\sum_{i=N/2+1}^{N} I_i^2 \right)   - \frac{N}{2} \ln \left( 2 \pi \sigma_I \sigma_O \right).
\end{align}

\begin{figure}[t]
\centerline{\includegraphics[height=7cm]{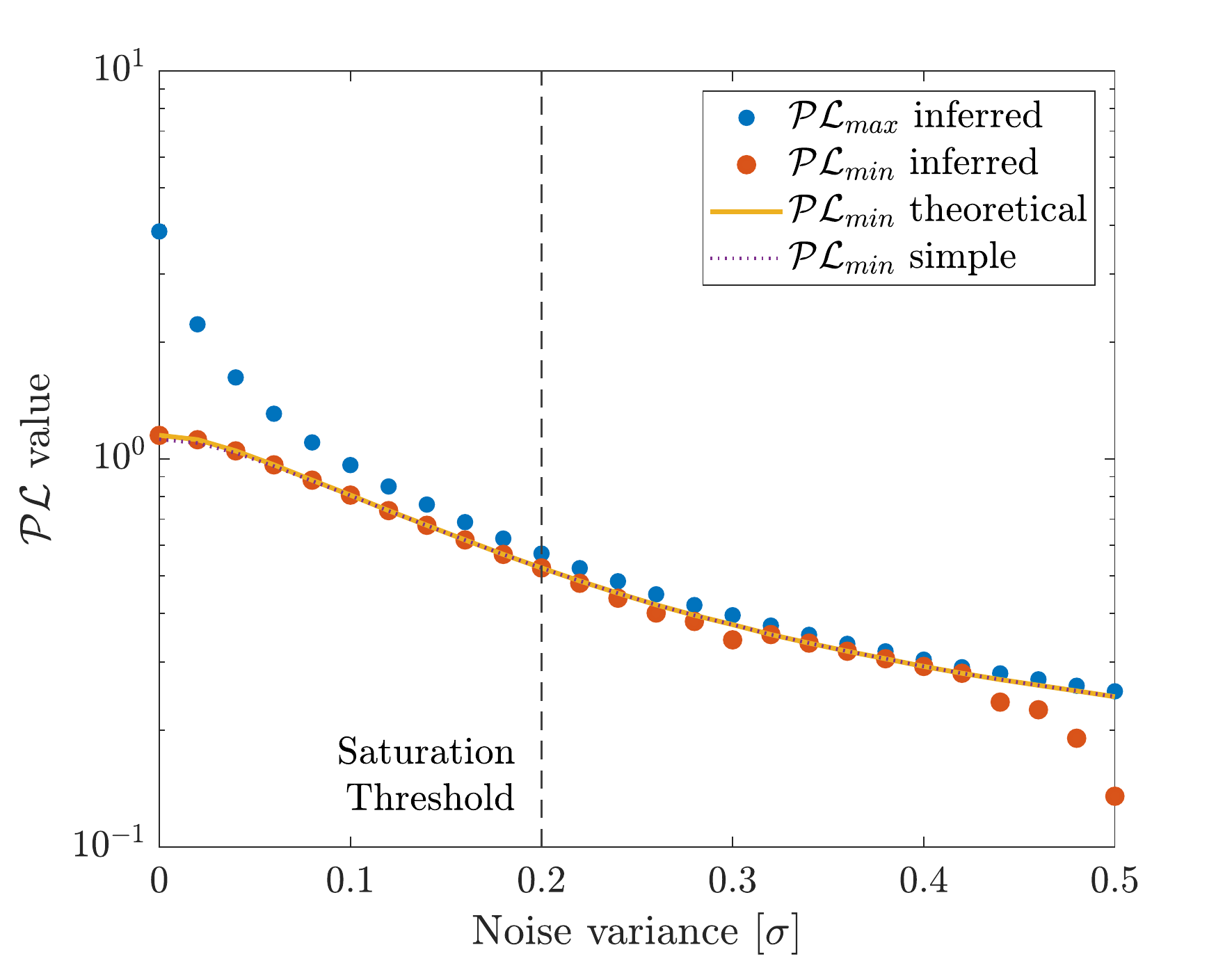}}
\caption{Pseudolikelihood boundaries as a function of the noise plotted in a log-scale. The blue values are the $\mathcal{L}$ minimized with all the parameters, that progressively decimate to the final orange values. The continous yellow and dashed violet line represents the values expected from the theory, respectively from equations (\ref{eq:PL_min}) and (\ref{eq:PL_min_simple}).}
\label{fig:Fig3_PLminmax}
\end{figure}

For example in the case of 4x4 of Fig. \ref{fig:Fig2_IOstatistics}, we have $\sigma_I = \sigma^{\rm in}$ and $\sigma_O = 0.05$. In the following, we will consider the results obtained by a uniform shift of the intensities $\bm I- 0.5$, thanks to the statistical distribution properties of both input and output (Fig. \ref{fig:Fig2_IOstatistics}). If we consider this specific case, it is possible to calculate the TIC theoretically for the empty model. In Fig. \ref{fig:Fig3_PLminmax} we plot the numerically optimized minimum and maximum value for the $\mathcal{L}$, against the theoretical results. The plot clearly tells that the eq. \ref{eq:PL_min}-\ref{eq:PL_min_simple} predicts the lower boundary for the pseudolikelihood.

\section{Results}
\label{sec:results}
In the previous chapters, we have extensively described the learning framework proposed for an I/O spin-glass model. In the following we present the results found under different operative conditions, in order to define standard modus operandi. We will always keep an eye on theory, to be able to interpret the results. In order to ease the graphical visualization of the results, we focus on the problem of 4x4 I/O matrix. Such results will be directly extendible to any I/O size, taking into account the possible presence of undersampling problematics that we will examine extensively.

\subsection{Inferring $\mathbb M$ with all the parameters}
\begin{figure}[t!]
\centerline{\includegraphics[height=7cm]{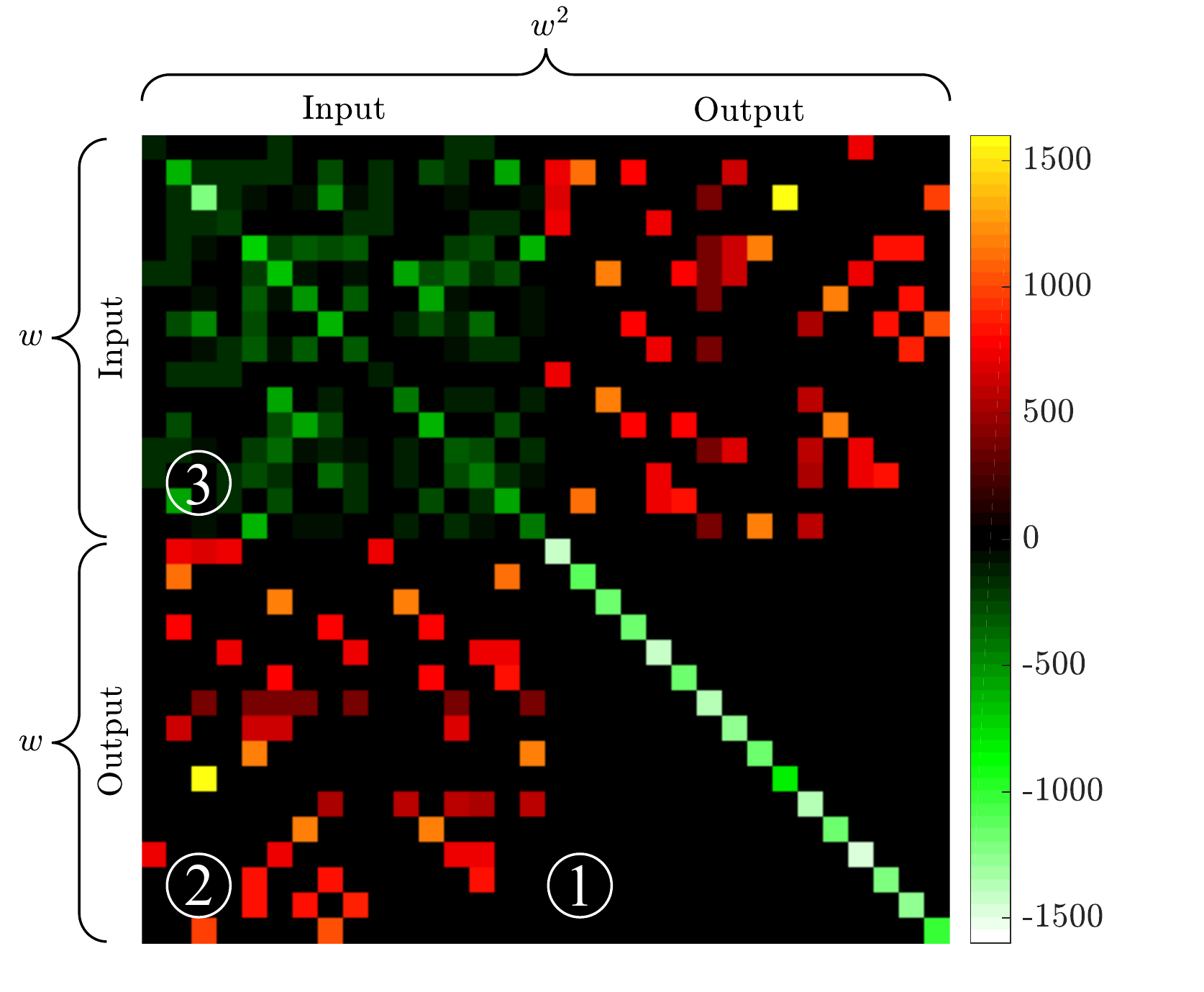}}
\caption{Schematics of the initial inference results: the matrix $\mathbb M$ obtained after the $\mathcal{L}$ maximization on independent active parameters. The black regions are the values close to zero, while with green and red we can distinguish between three (respectively negative and positive) different matrix subregions.  }
\label{fig:Fig05_matrixJ}
\end{figure}

To begin, let us focus on the algorithm convergence. We keep all the parameters free to adjust themselves 
into a form that maximizes the pseudolikelihood without decimation. Here, we explicitly consider the inference results of a 4x4 I/O system, implying a $\mathbb M$ matrix of 32x32 parameters. Similar results were found with every size considered, but for the sake of a clear visualization we initially present the results of the smallest system. 
We consider the minimization with weak noise $\sigma=0.02$ and a positive $\mathbb{T}$ matrix with $s=0.2$ sparsity. After maximizing the pseudolikelihood with respect to $M_{i\leq j}$, we obtain the result shown in Fig. \ref{fig:Fig05_matrixJ}. 

Let us call the inferred matrix
 $\mathbb{M}^{\rm inf}$.
 It is quite clear that the matrix converged into the expected tensorial form given by (\ref{def:M}):
  the green area is the negative part of the matrix, while in red it is represented the positive. We label three parts in the matrix, as shown in the figure:
\begin{list}{•}{}
\item Part 1) - The self-output coupling matrix converges into a diagonal form, $-\bm{\beta}$, whose elements are  $\{\beta_{i-N/2}\}=(\beta_1, \beta_2, \ldots, \beta_{N/2})$, for $i=N/2+1, \ldots, N$.
\item Part 2) - The input-output coupling matrix closely follows the transmission matrix $2 \bm{\beta}\mathbb{T}$ used for the dataset creation.
\item Part 3) - The self-input coupling matrix reported is  $-\mathbb{V}$, with, cf. Eq. (\ref{def:V}),
\begin{eqnarray}
\nonumber
-V_{\eta\xi}&=& - \sum_{\alpha=1}^{N/2}   T_{\eta\alpha}^\dagger \beta_{\alpha} T_{\alpha\xi}^{\phantom{\dagger}}   \quad 
\qquad \forall \eta,\xi=1,\ldots, N/2, \quad i\leq j \ ,
\\
-\mathbb{V}&=& - \mathbb{T}^{\dagger}{\mathbb{\bm \beta}}\mathbb{T}
\label{eq:betaU}
\end{eqnarray}
\end{list}
 To better understand how the noise propagates within this matrix, it is convenient to separate the formal analysis 
 per each of the parts labelled in the interaction matrix $\mathbb{M}$, 
 separating out   the noise mask $\bm{\beta}$ from matrix (\ref{def:M}).
 
\begin{figure}[t!]
\centerline{\includegraphics[width=14cm]{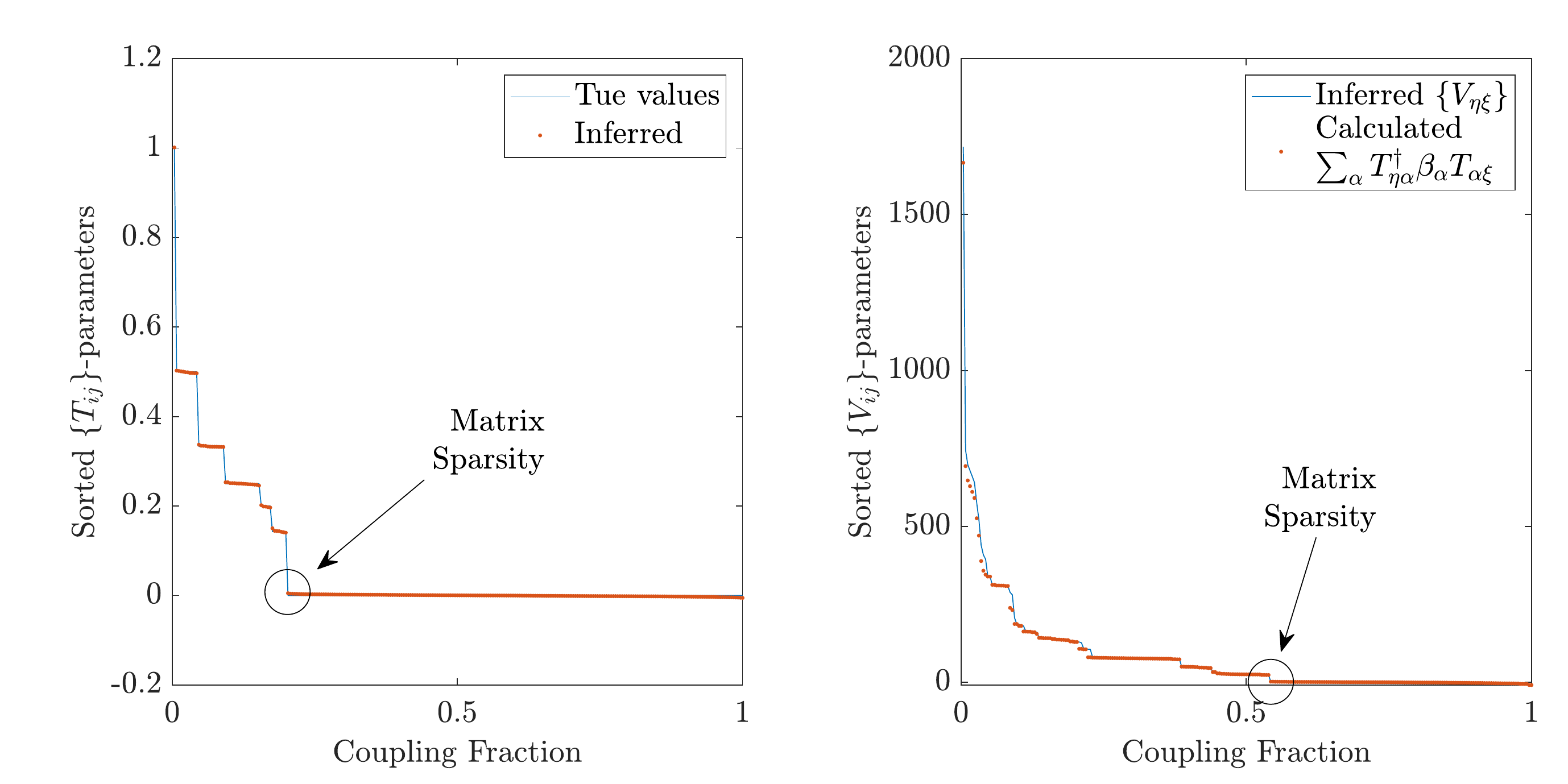}}
\caption{Part a), in orange the sorted values of $\mathbb{T}_{\rm inf}$ against the true values $\mathbb{T}$ (blue line). Part b), sorted values of $\mathbb{V}_{\rm inf}$ against the $\mathbb{V}$. The plots of parts a-b) are obtained scaling the values in the matrix of part a) with the rules given in sections \ref{ssec:temperature_channel}, \ref{ssec:transmission_matrix} and \ref{ssec:u_matrix}.}
\label{fig:sortedTandV}
\end{figure}

\subsubsection{The noise channel as statistical temperature} \label{ssec:temperature_channel}
First of all, it is important to notice that the diagonal term (Part 1 of $\mathbb M^{\rm inf}$) gives us an estimation of the noise per each channel of the system. 
According to the tensorial form (\ref{def:M}), in the diagonal part we have that $M_{ii} = \beta_i$ for $i>N/2$, thus the diagonal elements of $\mathbb M^{\rm inf}$ explicitly recovers the noise $\beta_{\gamma=i-N/2}$ per each I/O channel as in eq. (\ref{eq:sep_hamiltonian}). It is interesting to see how the perturbative noise statistics can be recovered by looking directly at the diagonal part. We studied a range of noise $\sigma$, that we can relate to the factor $\beta_\gamma$ for each output channel $\gamma$, via:
$\beta_\gamma = 1/2/ \sigma_\gamma^2$, $ \gamma = 1, \ldots, N/2$, i. e., 
in matrix notation, 
\begin{equation}
\bm \beta = (2 \bm \sigma^2)^{-1}
\end{equation} 
where  the diagonal matrix $\bm \sigma$ is introduced in  Eq. (\ref{eq:transmission}).

\begin{figure}[b!]
\centerline{\includegraphics[height=7cm]{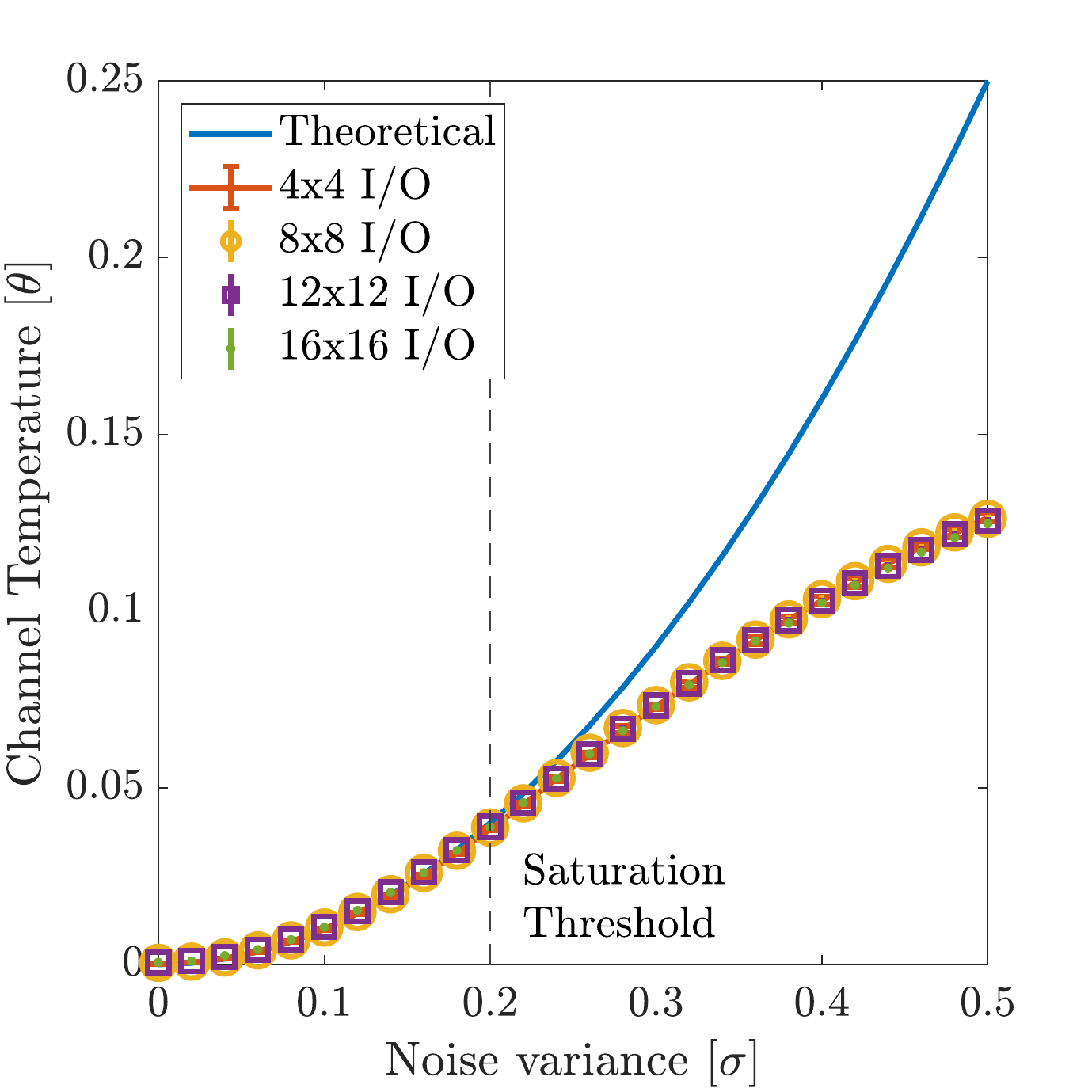}
\includegraphics[height=7cm]{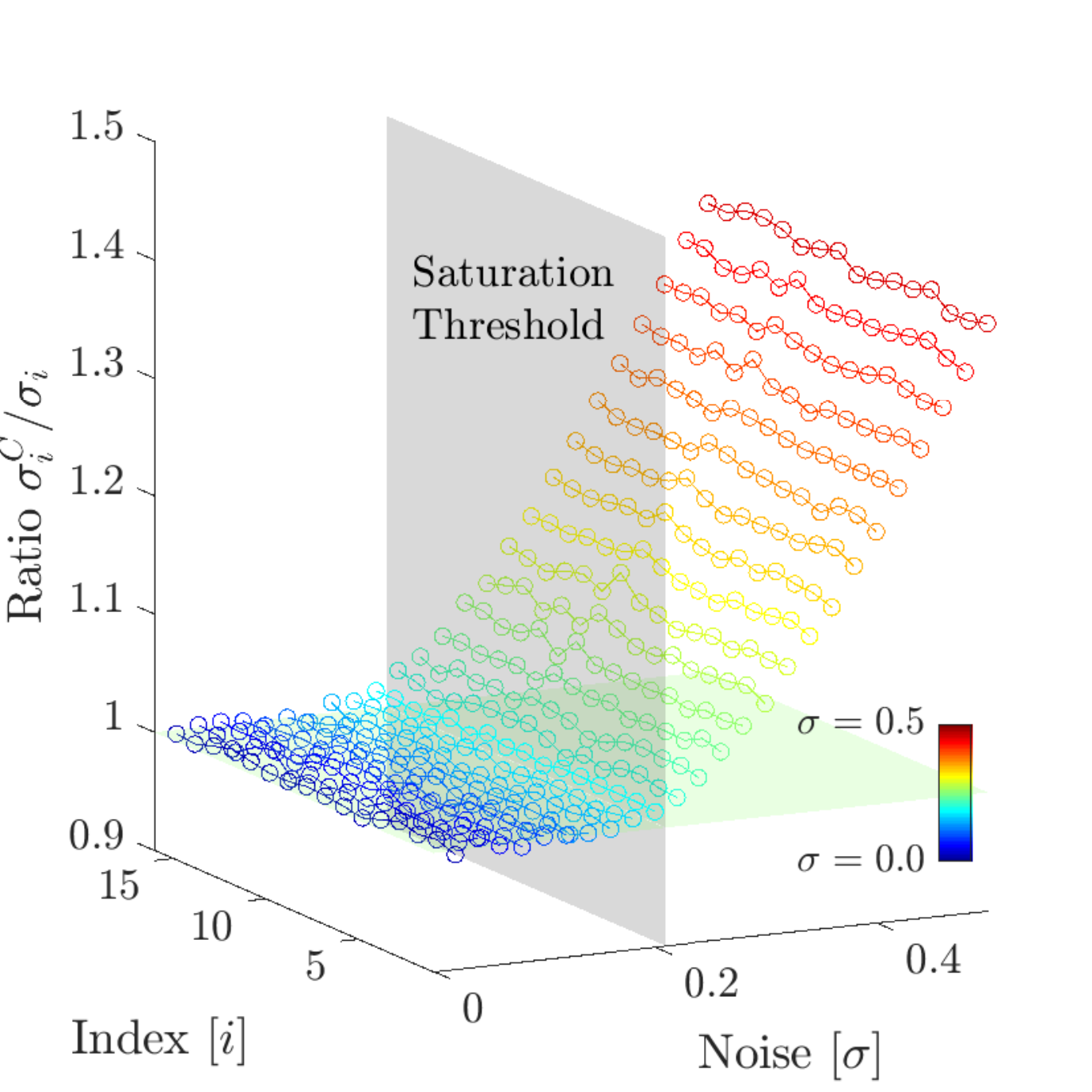}}
\caption{Part a), plot of the mean temperature $\theta$ obtained for the four systems tested. The theoretical value (blue dashed line) is calculated with $\theta_{\rm th} = 2\sigma^2$. Part b), plot of the ratio between the calculated variance of the dataset per each line $\sigma_i^C$ normalized by the $\sigma_i$ obtained in the diagonal part of matrix $\mathbb M$. The gray plane represents the expected ratio when both the estimation and the effective noise are scaling like the noise inserted in the system $\sigma$.}
\label{fig:temperatureAnalysis}
\end{figure}

Thus, giving a thermodynamical interpretation,  the average temperature of the measurement is:
\begin{equation}
\theta \equiv \frac{2}{N} \sum_{\gamma=1}^{N/2} \frac{1}{\beta_\gamma}
\end{equation}
\noindent
where, theoretically, the temperature can be expressed in terms of the variance of the noise introduced in Eqs. (\ref{eq:delta_lim}) or (\ref{def:betagamma}), $\theta_{th}= 2 \sigma^2$. In Fig. \ref{fig:temperatureAnalysis}, we plot the average temperatures, estimated from the inference of the full model for the systems studied, as a function of the noise inserted in the dataset. We can appreciate a pretty good agreement with the theoretical expectation, that is the system temperature scales quadratically with the inserted noise. After a certain point around $\sigma = 0.20$, the estimation departs from the parabolic trend. This effect is due to the effective statistics of the output intensity values in Fig. \ref{fig:Fig4_outputpdf_vs_noise}, where an important saturation effect emerges when the noise is high compared to the intensity range accepted by the system.

We can conclude that the diagonal matrix of part 1) is effectively giving an estimation for the  noise experienced within each 
transmission channel. This can be visualized in part b) plot of Fig. \ref{fig:temperatureAnalysis}, where per each channel we calculate the inserted noise standard deviation $\sigma_i^C$ and we divide it for the estimated one $\sigma_i$, as a function of the variance of the noise $\epsilon$ introduce in Eq. (\ref{eq:transmission}). It is possible to notice that this ratio stays around $1$ (the gray plane), indicating that we are effectively estimating the channel noise up to a perturbation of $20\%$. Beyond this value, the noise estimation starts to depart from the real one due to saturation effects.

\subsubsection{The Transmission matrix} \label{ssec:transmission_matrix}
Understanding the diagonal matrix in block (1) gives us the possibility to calculate exactly the transmission matrix elements $\mathbb T^{\rm inf}$ from the inferred $\mathbb{M}^{\rm inf}$ block (2): we divide row wise $M_{ij}^{\rm inf}$ for $i\in[1,N/2]$ and $j\in[N/2+1,N]$ 
by the elements $\beta_i$, so that the estimation of the system transmission matrix is given by
\begin{equation}
T^{\rm inf}_{i,j-N/2} = \frac{M^{\rm inf}_{ij}}{2\beta_i}.
\end{equation}  
Part a) of Fig. \ref{fig:sortedTandV} shows the sorted values against the matrix sparsity $s$ reconstructed at $\sigma=0.05$. We can clearly see how the calculated $T^{\rm inf}_{ij}$ values closely match the expected blue curve, abruptly falling at the exact matrix sparsity. However, at this stage also the other parameters are active and we are interested in cutting out all those which are irrelevant for the correct description of the $\mathbb{T}$ matrix.

\subsubsection{Self-Input coupling} \label{ssec:u_matrix}
Also Part 3) of $\mathbb M^{\rm inf}$ is closely entangled with the noise estimation, cf. Eq. (\ref{eq:betaU}). We formally expect $M_{ij}=M_{ji}$, and, as we already said, we implement the symmetry in the minimization procedure.
For this part,
we compare
 $\mathbb{V}^{\rm inf}$ with $\mathbb{T}^\dagger \bm\beta^{\rm inf} \mathbb{T}$, where the $\mathbb{T}$ is the true transmission matrix used to produce the dataset.
Part b) of Fig. \ref{fig:sortedTandV} reveals the appropriateness of the values recovered for the $\mathbb{V}^{\rm inf}$ against the theoretically expected trend, with the inferred set of output channel noises $\bm{\beta}^{\rm inf}$.

\subsection{Decimation and Information Criterion comparison} \label{sec:directT}
The results presented above are referred to the best model learned with all the parameters active in $\mathbb{M}$. However, this is not always the best choice due to possible overfitting problematics: using all the parameters will fit well the training dataset, but it will poorly perform on the effective estimation of the transmission matrix representing the system, eventually failing the fit procedure on the validation dataset. To avoid this, we are interested in estimating the exact number of parameters active in the channel. We, therefore, proceed as proposed in the decimation procedure, in which we recursively eliminate the smallest couplings, in successive $\mathcal{L}$-maximization processes. While, doing this, we expect an abrupt change in the $\mathcal{L}$ value at the over- to under-fitting connectivity threshold, which will point out the best model fitting the dataset. 

The quality of the inference procedure depends on the sampling rate, Eq. (\ref{def:samp_rate}), that weights the number of measurements against the number of parameters to be estimated in the model, thus we study the behaviour of our inference method under a variety of numerical conditions. 
As a function of the linear input size $w$ it is 
\begin{equation}
\xi \left(w\right) = \frac{M}{K(w)}.
\end{equation}
By construction the matrix $\mathbb M$ has $K(w) = 4w^4$ entries, however taking into account that such matrix is symmetric and that 
the lower diagonal block in Eq. (\ref{eq:generalizedJ}) is diagonal, the number of total independent parameters to be estimated is given by $K(w)= 3/2 \left(w^4 +w^2 \right)$.
Furthermore, even though we will always start under the assumption of largest ignorance, eventually, in the case of the direct transmission matrix inference, this will only have a fraction $s$ of non-zero elements. Therefore the total number of parameters drops to  $K= (s+1/2) w^4+3/2 w^2$.
 In  the following, considering $M=10000$ I/O couples, $s=0.2$ and varying $w$, we find the sampling ratio values reported in Tab. \ref{tab:xi} both for the direct (sparse) case and for the inverse (complete) transmission matrix.
 In both cases 
  the four datasets span a complete sampling scenario, from best to worst.

\begin{table}[t!]
\begin{center}
\begin{tabular}{|c|c|c|}
\hline
$w\times w$ &$\xi (s=0.2)$ &$\xi$ (complete) \\
\hline
    $4\times 4$  &$49.21$ & $24.5$ \\
     $8\times 8$  &$3.37$&$1.60$\\
     $12\times 12$  &$0.68$&$0.32$\\
     $16\times 16$  &$0.22$&$0.10$ \\
     \hline
\label{tab:xi}
\end{tabular}
\caption{Sampling ratio for direct (diluted) and inverse (complete) effective transmission matrices for $M=10000$.}
\end{center}
\end{table}

\begin{figure}[b!]
\centerline{\includegraphics[height=7cm]{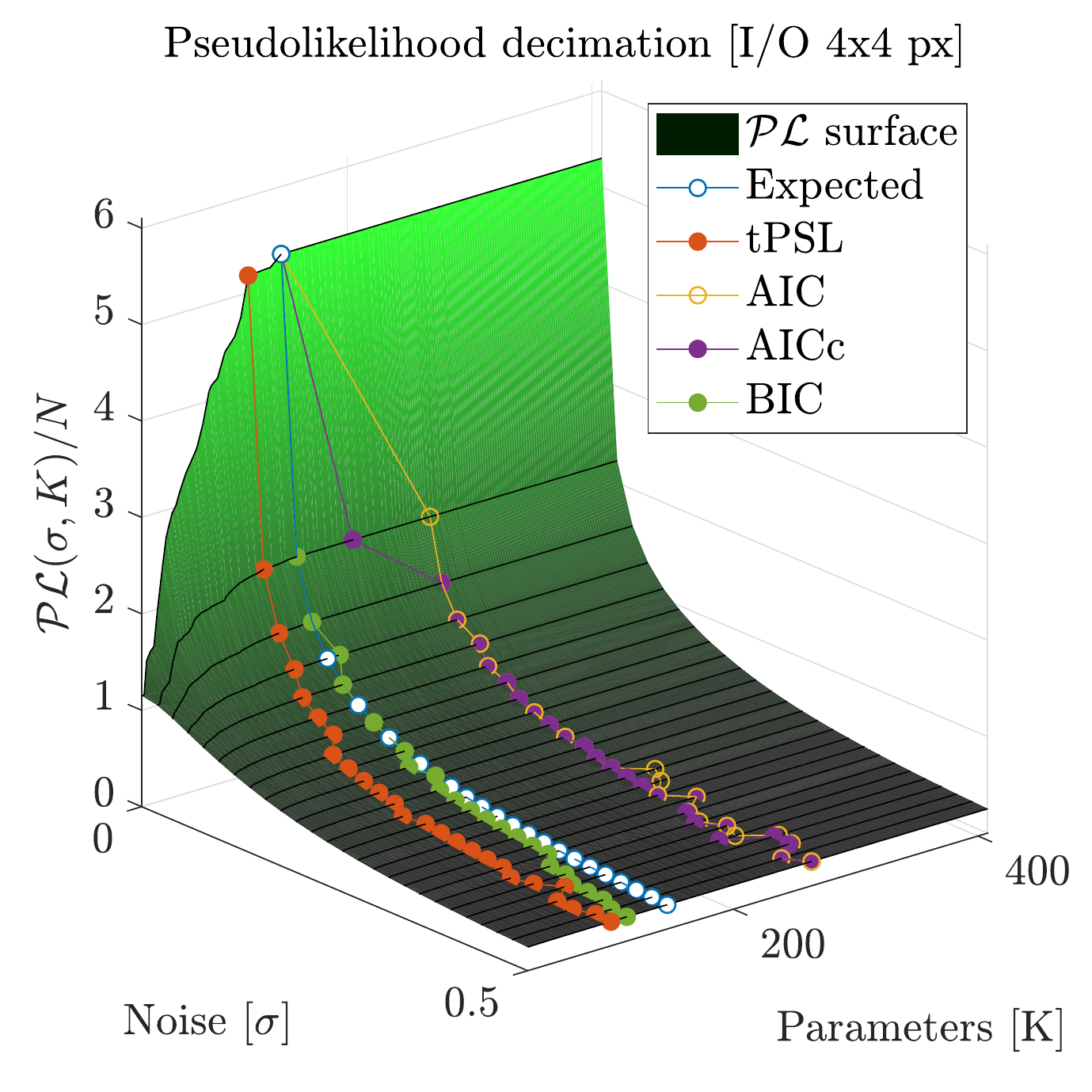}
\includegraphics[height=7cm]{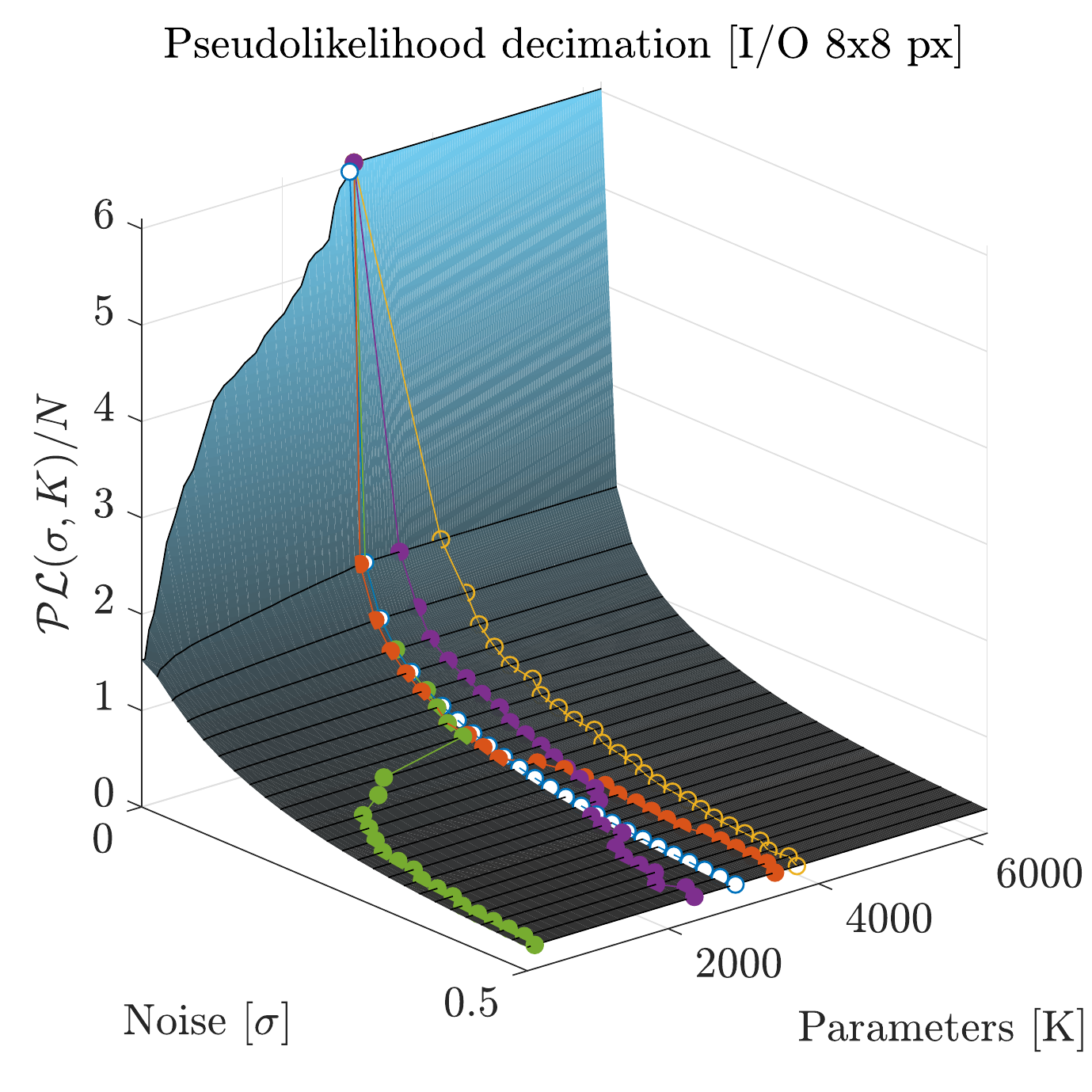}}
\centerline{\includegraphics[height=7cm]{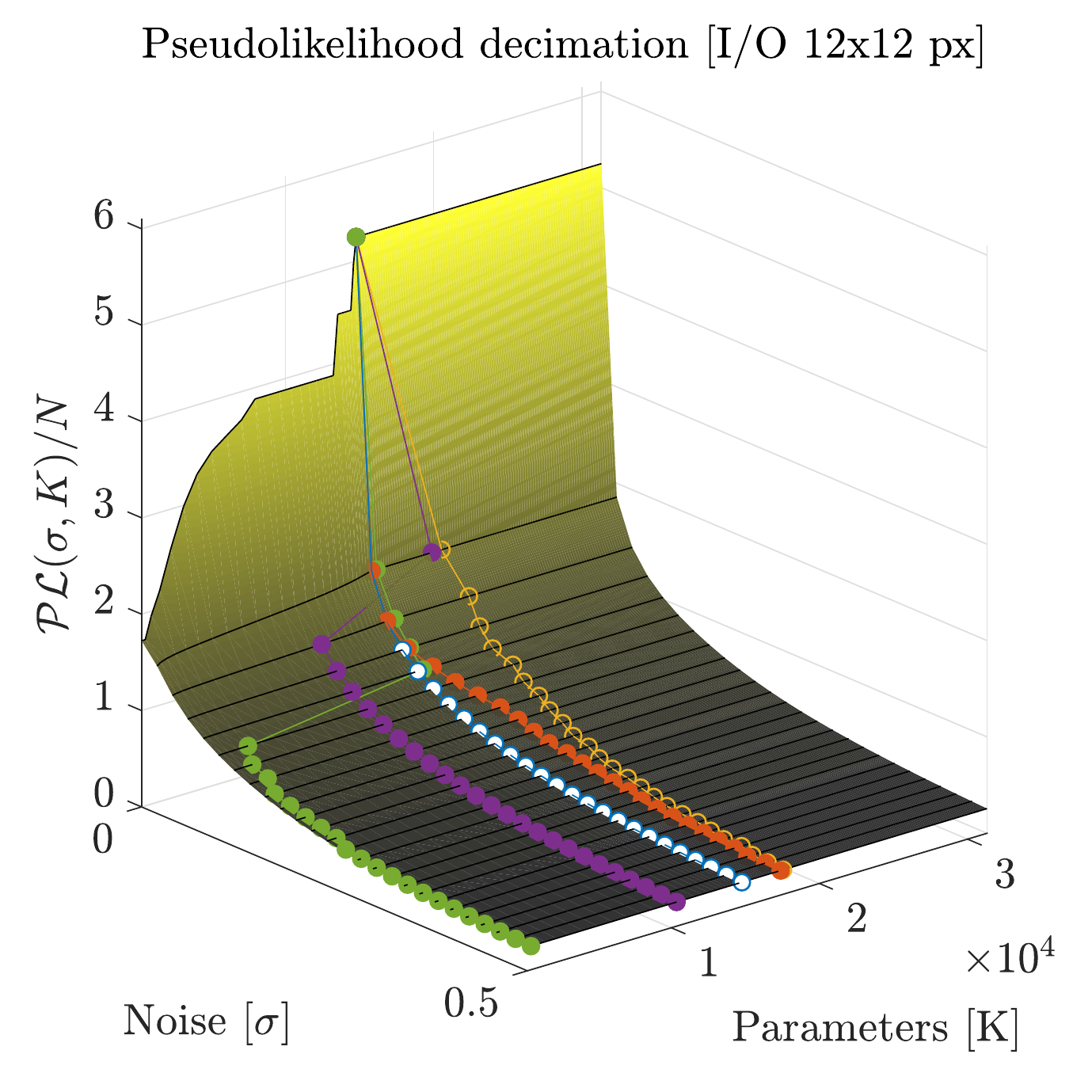}
\includegraphics[height=7cm]{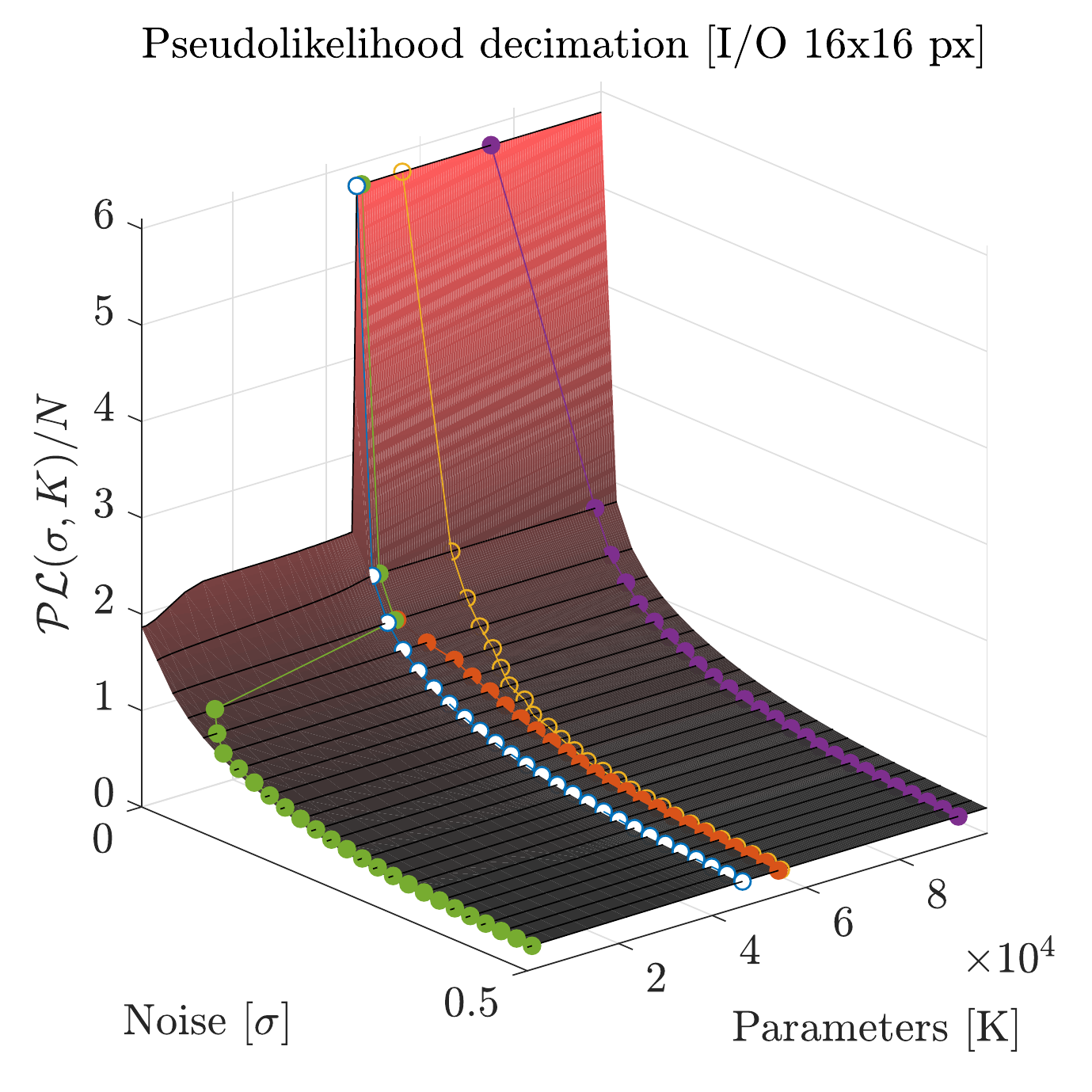}}
\caption{$\mathcal{L}$-surface plots of four I/O systems tested in this study, differring in the input size $w$. The surface represent the pseudolikelihood optimized value as a function of the noise and the parameter number (thus, the decimation).}
\label{fig:Fig08_PLsurf}
\end{figure}

We, further, run independent optimizations as a function of the noise $\sigma$ and study the behaviour of the $\mathcal{L}$ progressively eliminating a fraction of $1/128$ couplings at each decimation step. To pick the optimal model among the decimation procedure, we evaluate the information criterions described in Section \ref{ssec:decimationstrategy}, and we compare them on the $\mathcal{L}$-surface plots in Fig. \ref{fig:Fig08_PLsurf}. In each plot, the exact number of parameters to be inferred is represented with a blue line with white dots.
Per each sampling rate, we encounter different behaviours, according to the regimes sketched in Sec. \ref{sec:MeaSam}:

\begin{list}{•}{}
\item \textbf{I/O 4x4} ($\xi \gg 1$): this is the best case for Bayesian inference, all the IC correctly guess the model number at zero noise $\sigma=0$. As soon as the noise is inserted, the situation changes. AIC and AICc follow the same trend (this is expected since AICc converges to AIC in case of high sampling rate). Both result quite prudent, preferring always models with higher number of parameters. On the other hand, TIC excessively decimates at all the noises. BIC greatly outperfoms all the information criteria in this scenario, correctly guessing the parameter number up to $\sigma=0.36$.

\item I/O \textbf{8x8} ($\xi > 1$): the situation changes with respect the previous one. AIC is always very prudent and very stable against noise, while AICc pushes more toward the proper result. However at high noise, AICc excessively decimates. TIC instead, obscillates around the correct value, overdecimating at low noise and underdecimating at higher $\sigma$ values. BIC correctly guesses the parameter number up to a moderate noise, after that, BIC transits towards excessive decimation. BIC is really not effective for $\sigma\geq 0.16$. Noticeably, at low noise, TIC and BIC closely guess the number of parameters.

\item I/O \textbf{12x12} ($\xi \lesssim 1$): the previous behaviour changes in this sampling scenario. AIC is always the most prudent while AICc quickly overdecimates. TIC is very robust at this sampling rate, resulting always very close to the correct solution and never overdecimate. BIC is good up to $10\%$ noise, after that if fails, picking a practically empty model as  best guess.

\item I/O \textbf{16x16} ($\xi \ll 1$): In this regime, BIC fails at lower noises, recovering exact solutions up to a perturbation of $4\%$. TIC performs very well, staying prudent at higer noise and converging to same results as AIC. Remarkably, AICc tends to overfit, favouring results with higher number of active parameters.
\end{list}

\begin{figure}[h!]
\centerline{\includegraphics[width=7cm]{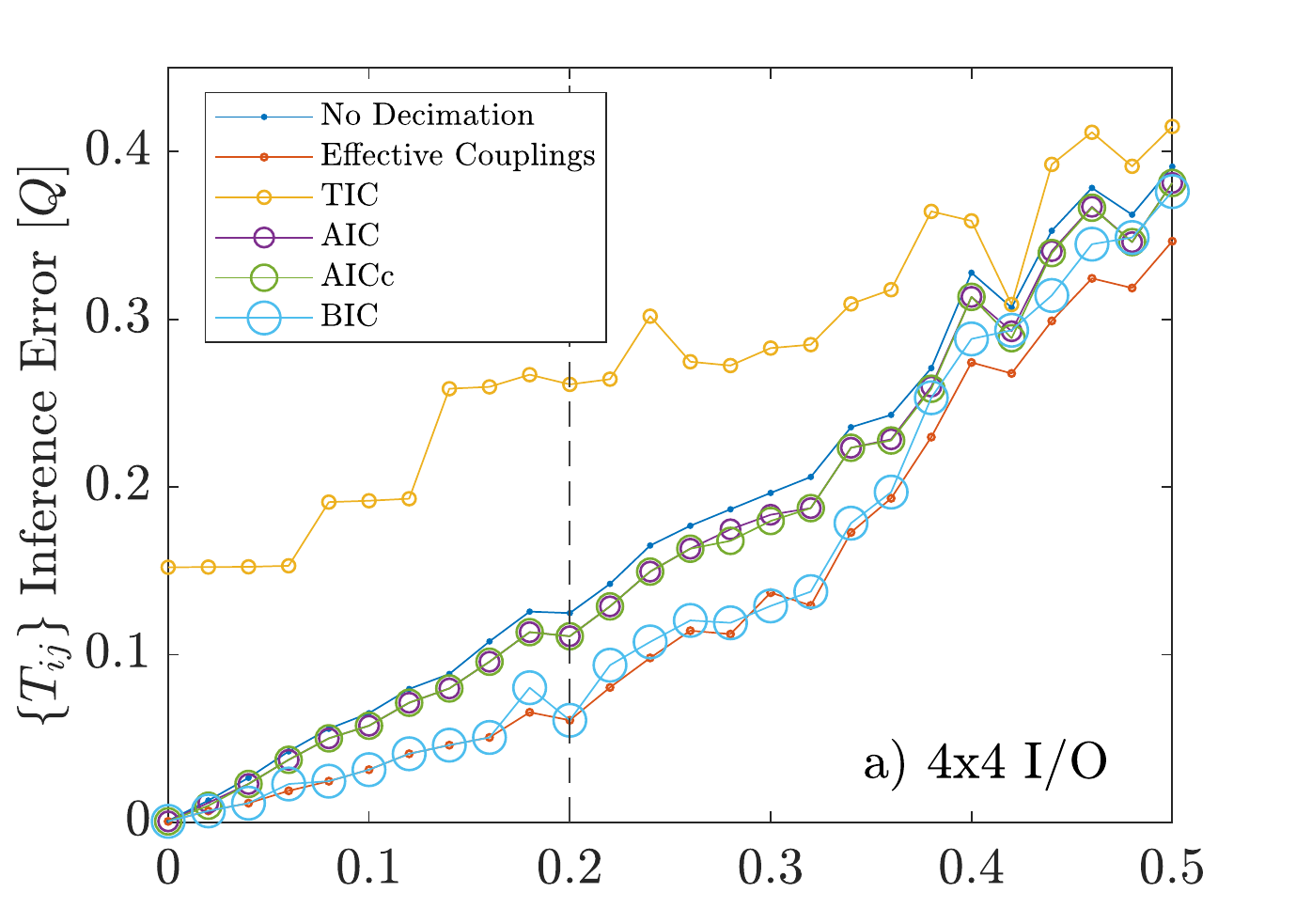}
\includegraphics[width=7cm]{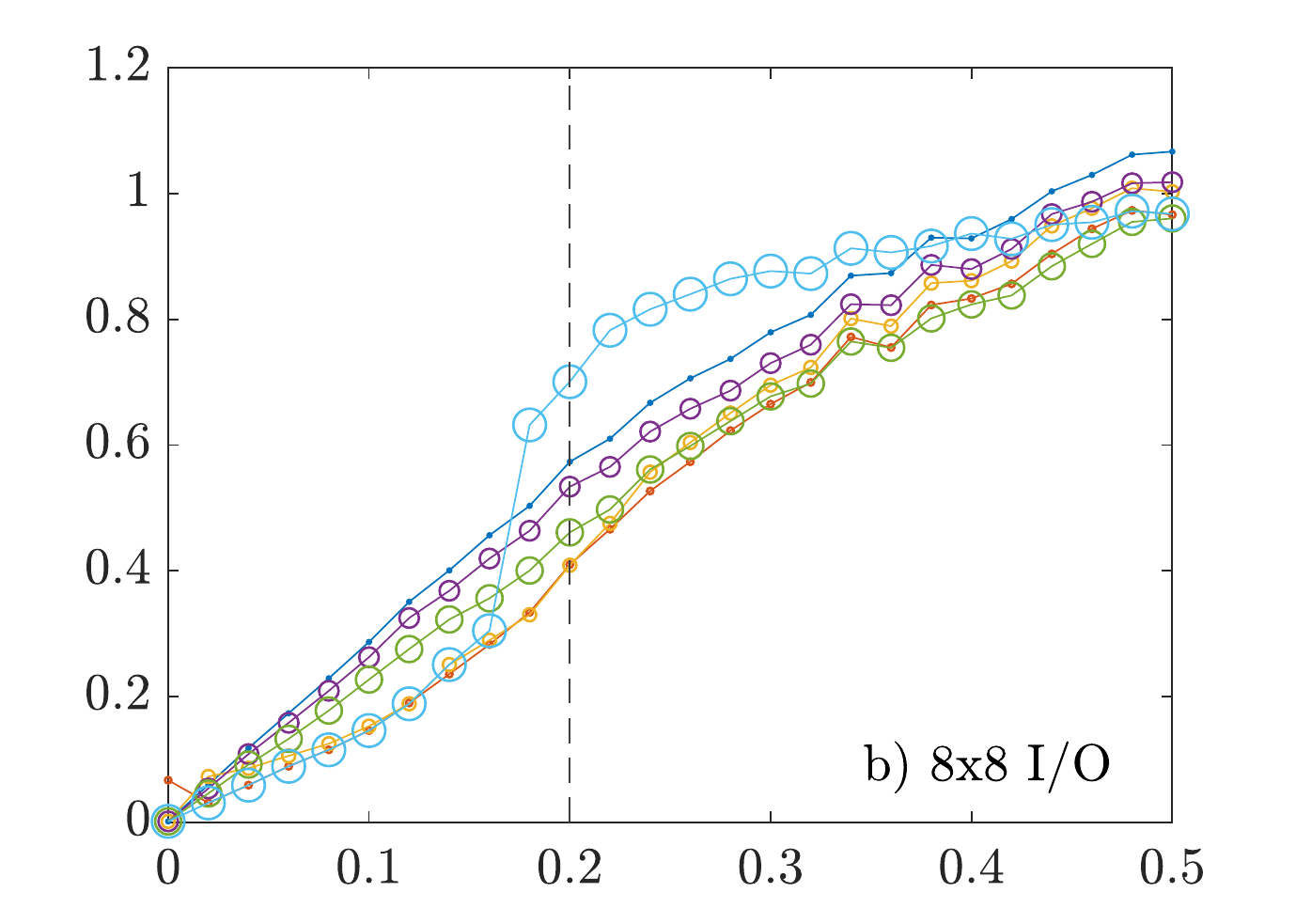}}
\centerline{\includegraphics[width=7cm]{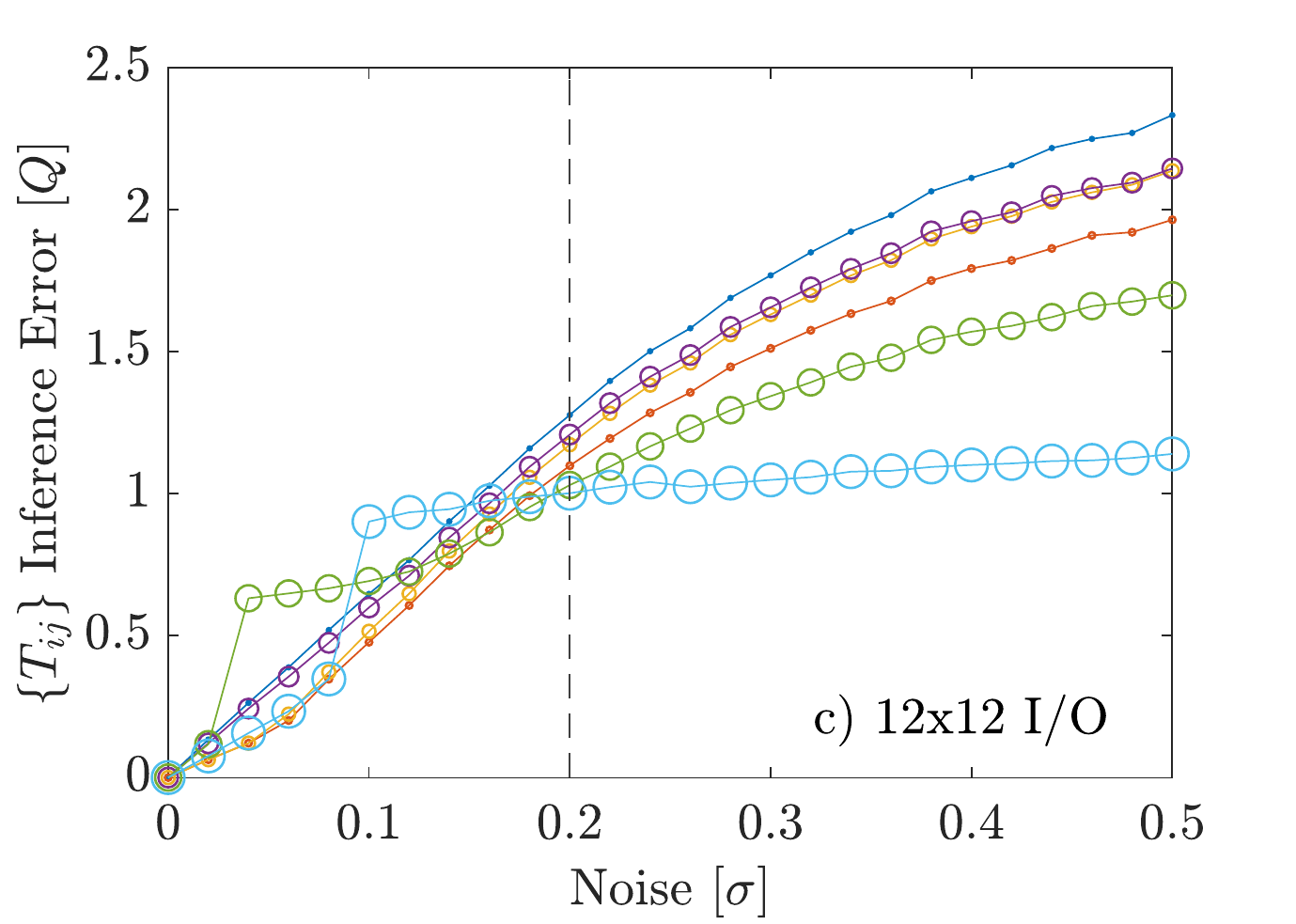}
\includegraphics[width=7cm]{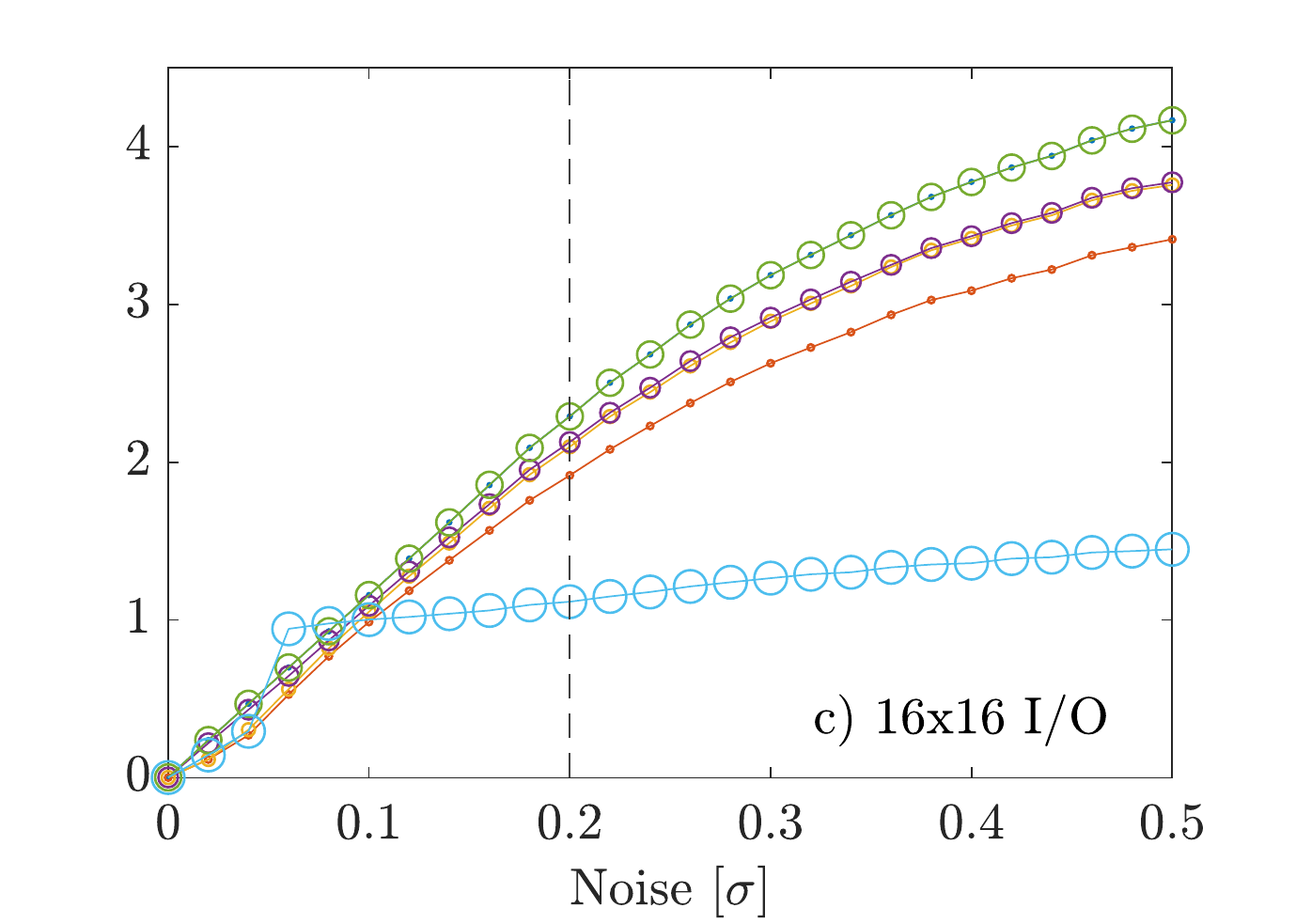}}
\caption{Reconstruction error $Q$ of the direct transmission matrix $\mathbb T_{\rm inf}$. The plots report the quantity $Q$ under different inference scenarios. We can appreciate that all the curves start from $Q\approx0$, growing as the noise and the size increase (with fixed number of measurements).}
\label{fig:T_reconstruction_quality}
\end{figure}

We have analyzed how ICs help finding the most accurate number of parameters among the decimated models quantitatively by defining the inference reconstruction error of the $\mathbb{T}_{\rm inf}$ against the original one (used to produce the datasets):
\begin{equation}
Q \left( \mathbb T, \mathbb T_{\rm inf} \right) = \sqrt{\frac{\norm{\mathbb T- \mathbb T_{\rm inf}}}{\norm{\mathbb{T}}}}.
\end{equation}

Here, $Q=0$ corresponds to exact recovery of $\mathbb{T_{\rm inf}}$ and in general low $Q$ represents good inference results. Results are presented in Fig. \ref{fig:T_reconstruction_quality} for all the ICs and model sizes.
Remarkably, at low noise we are able get excellent estimation of $\mathbb T$ and at all the sampling rates investigated. Although some IC perform better than others in picking the closest number of parameters, all the models are operatively similar in terms of absolute resutls. 
With a few exceptions: TIC poorly performs on oversampled systems (here at 4x4) at any noise and BIC progressively fails at lower noises as $\xi$ decreases. These facts are consistent with the plots of Fig. \ref{fig:Fig08_PLsurf}, giving a further indication in favor of an operative equivalence among the different selection criteria tested. In particular, no decimation resulted always worse than the others (thus appropriate decimation worth better results) and knowing the exact number of effective couplings decreases the $Q$-reference curve.

\subsection{The inverse transmission matrix} \label{sec:inverseT}
As already discussed, the same protocol can be used to infer the inverse transmission matrix $\mathbb{T}^{-1}$, rather than inferring $\mathbb{T}$ and then inverting it. This procedure will be used for the reconstruction of an image pattern that travels from the I-edge to the O-edge, thus to unlock the possibility to use an optically disordered media as a normal objective lens.
Let us analyze the reliability on the inference of the inverse, recalling that the dataset was generated with the direct $\mathbb T$ matrix. We point out that, in general, the $\mathbb{T}^{-1}$ does not have the same sparsity as the corresponding direct matrix, yet it is a stochastic matrix, as well. We proceed inferring inverse matrices increasing the noise level and the decimation. To be consistent, we present the results in analogy with section 
\ref{sec:directT}, using the same datasets and colorcode. 

\begin{figure}[t!]
\centerline{\includegraphics[height=7cm]{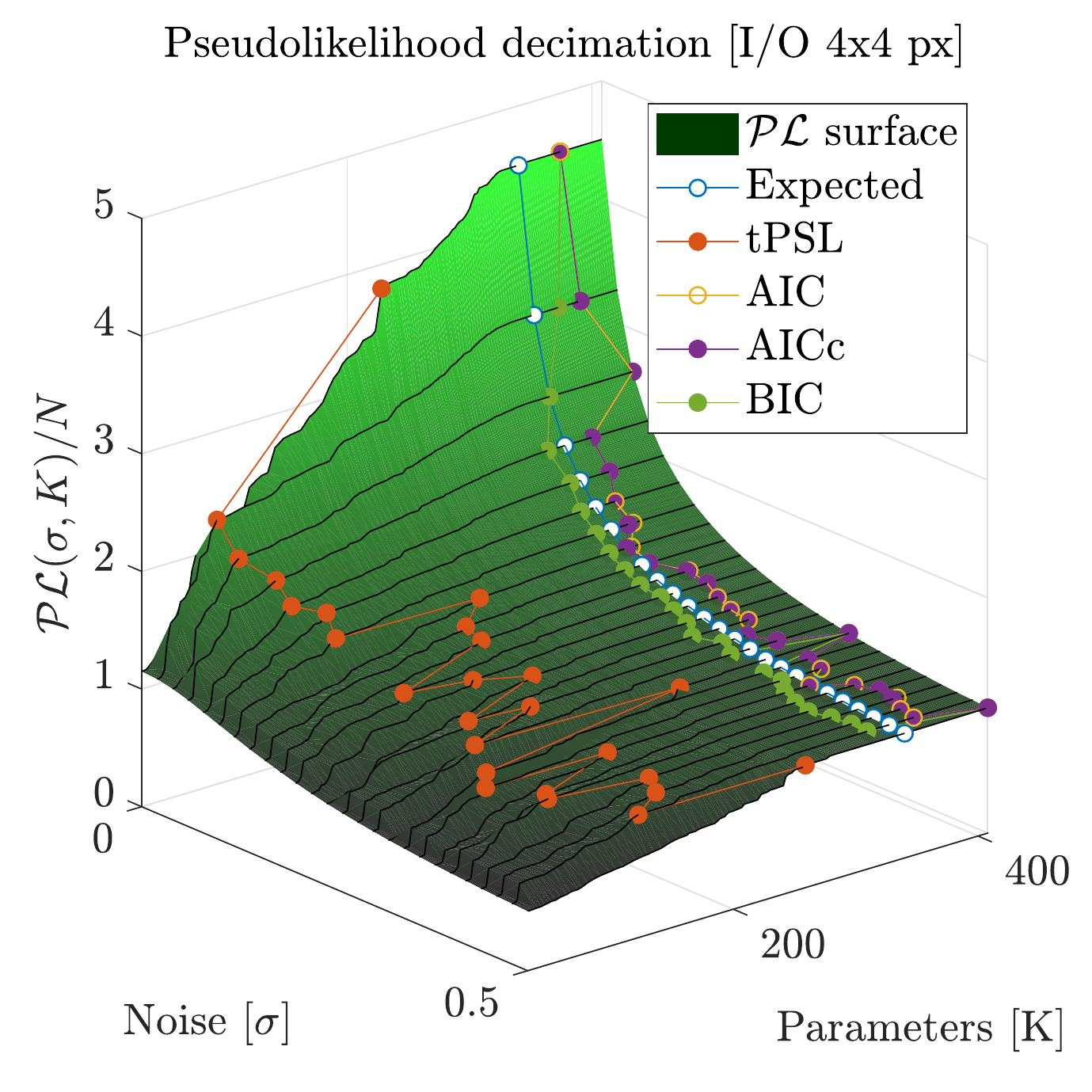}
\includegraphics[height=7cm]{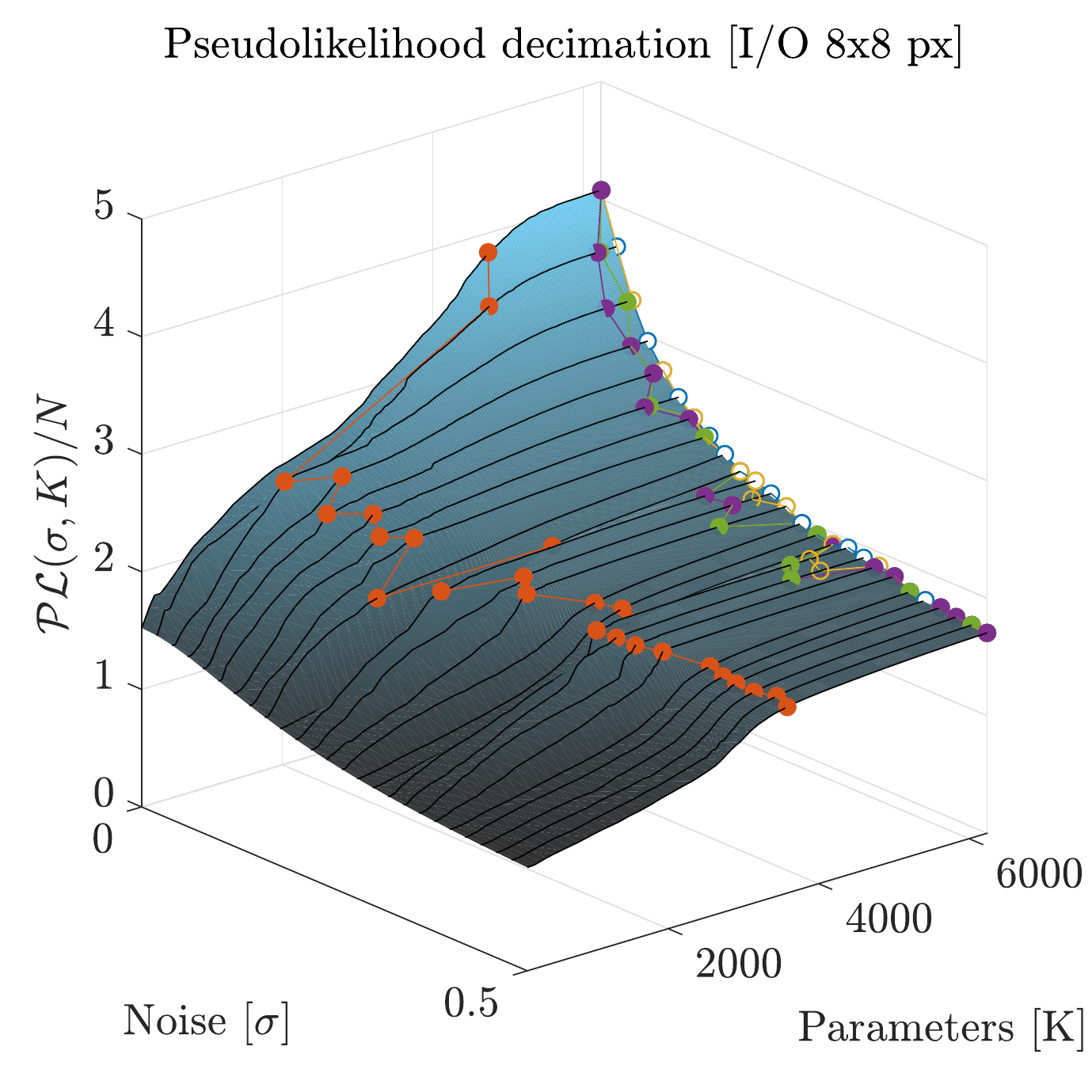}}
\centerline{\includegraphics[height=7cm]{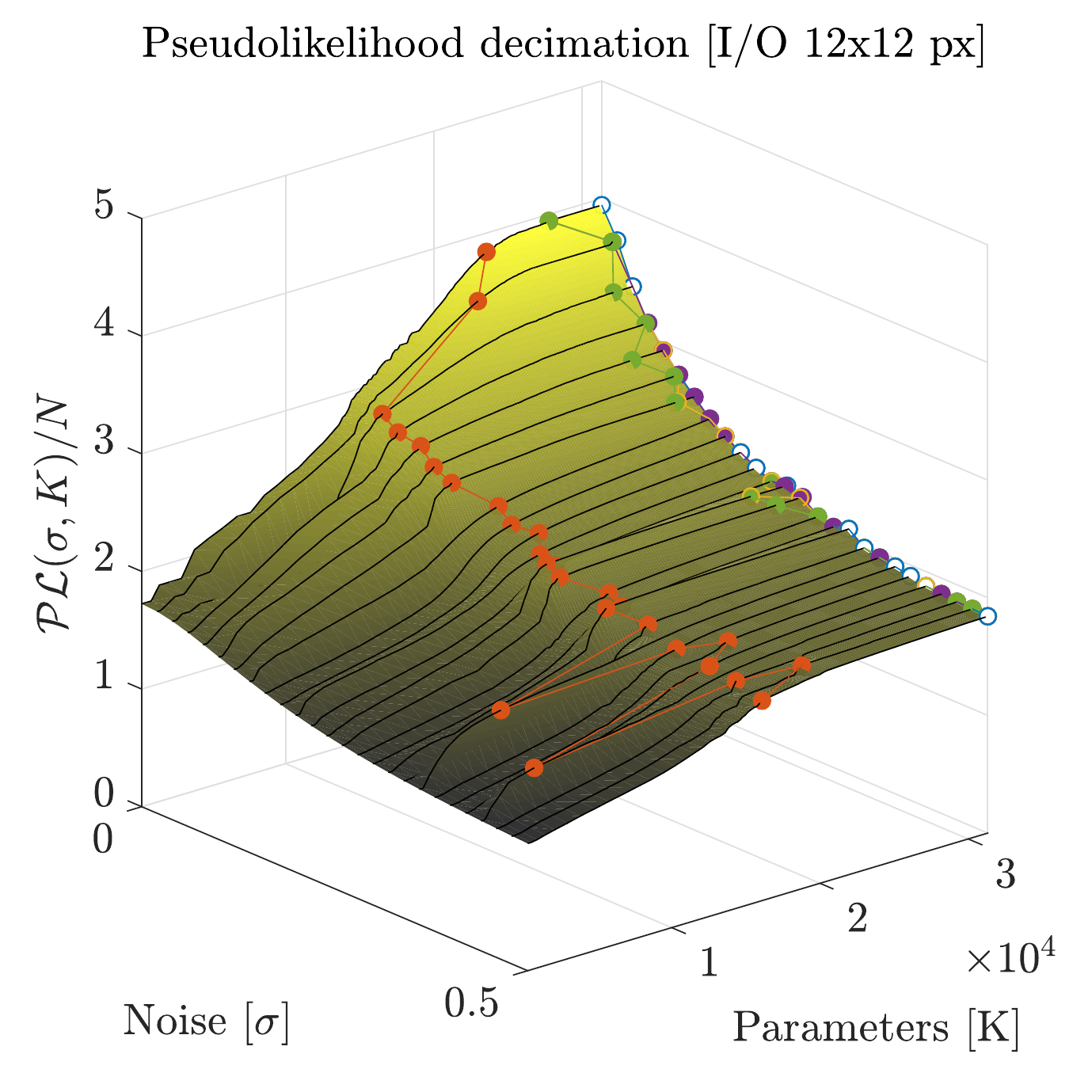}
\includegraphics[height=7cm]{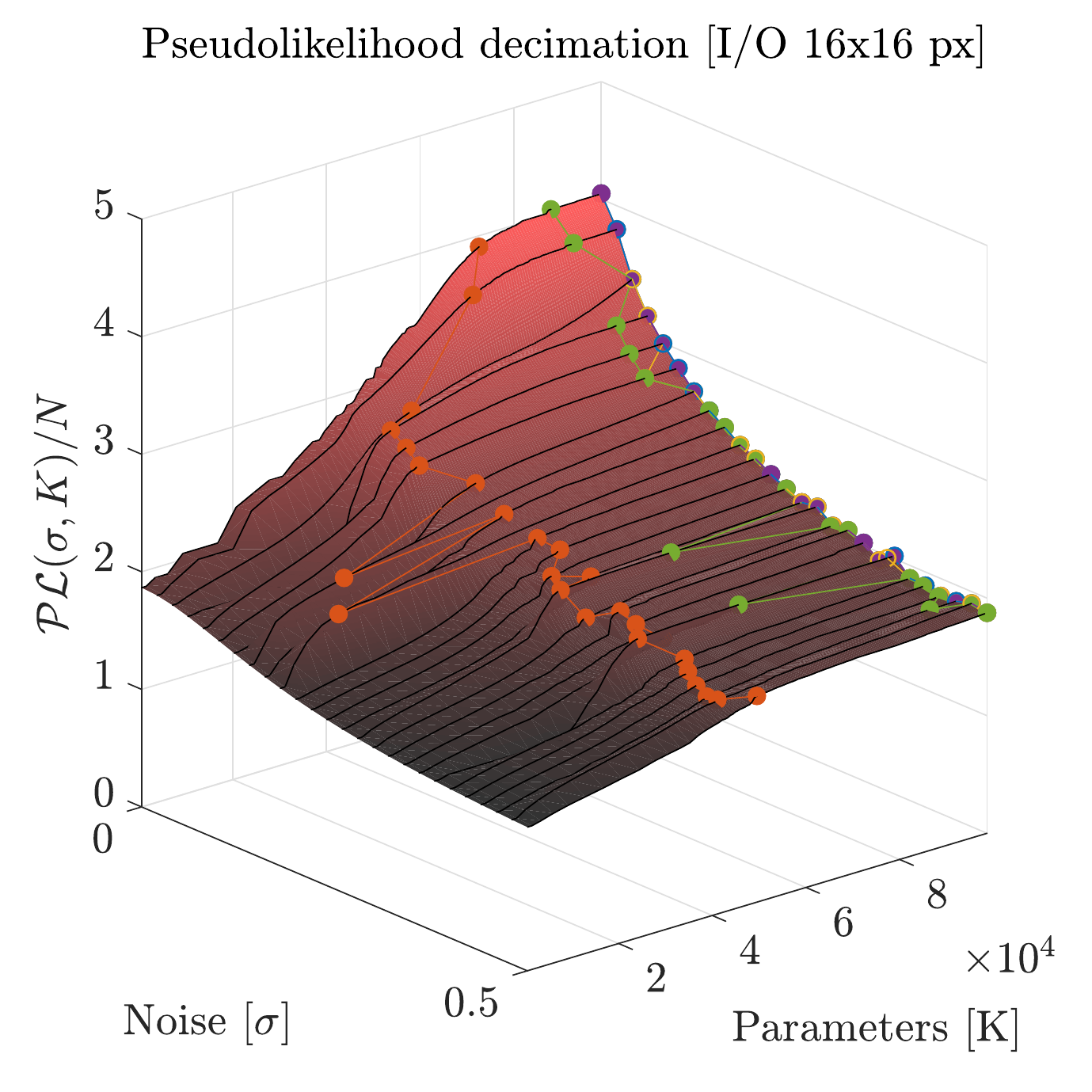}}
\caption{$\mathcal{L}$-surface plots of several I/O systems tested in this study differring in the input size $w$. The surface represent the pseudolikelihood optimized value as a function of the noise and the parameter number (thus, the decimation).}
\label{fig:inv_PLsurf}
\end{figure}

\begin{list}{•}{}
\item \textbf{I/O 4x4} ($\xi \gg 1$): this is the only size at which the inverse matrix has some null entries. Thus, there is a small degree of diluteness to be recovered. In the green $\mathcal{L}$-surface AIC and AICc are prudent with the decimation, BIC gets closer to the actual value of parameters to be estimated sligthly preferring setting to zero smaller couplings. Interestingly, TIC strongly decimates and has a less predictable behaviour, with strong fluctuation at high noise.

\item I/O \textbf{8x8} ($\xi \gtrsim 1$): at this size the $\mathbb{T}^{-1}$ is not sparse at all and, accordingly, AIC, AICc and BIC always determine a weak or no decimation. Like in the previous case, TIC tends to overdecimate the coupling matrix.

\item I/O \textbf{12x12} ($\xi < 1$): In this case, the situation is pretty similar to the 8x8 case, being TIC the worse information criterion among the others.
\item I/O \textbf{16x16} ($\xi \ll 1$): With the biggest system considered BIC tends to excessively decimate compared with AIC and AICc, which are the most accurate selection criteria. TIC fails like the previous cases, returning always a quite sparse matrix.
\end{list}

\begin{figure}[t!]
\centerline{\includegraphics[width=7cm]{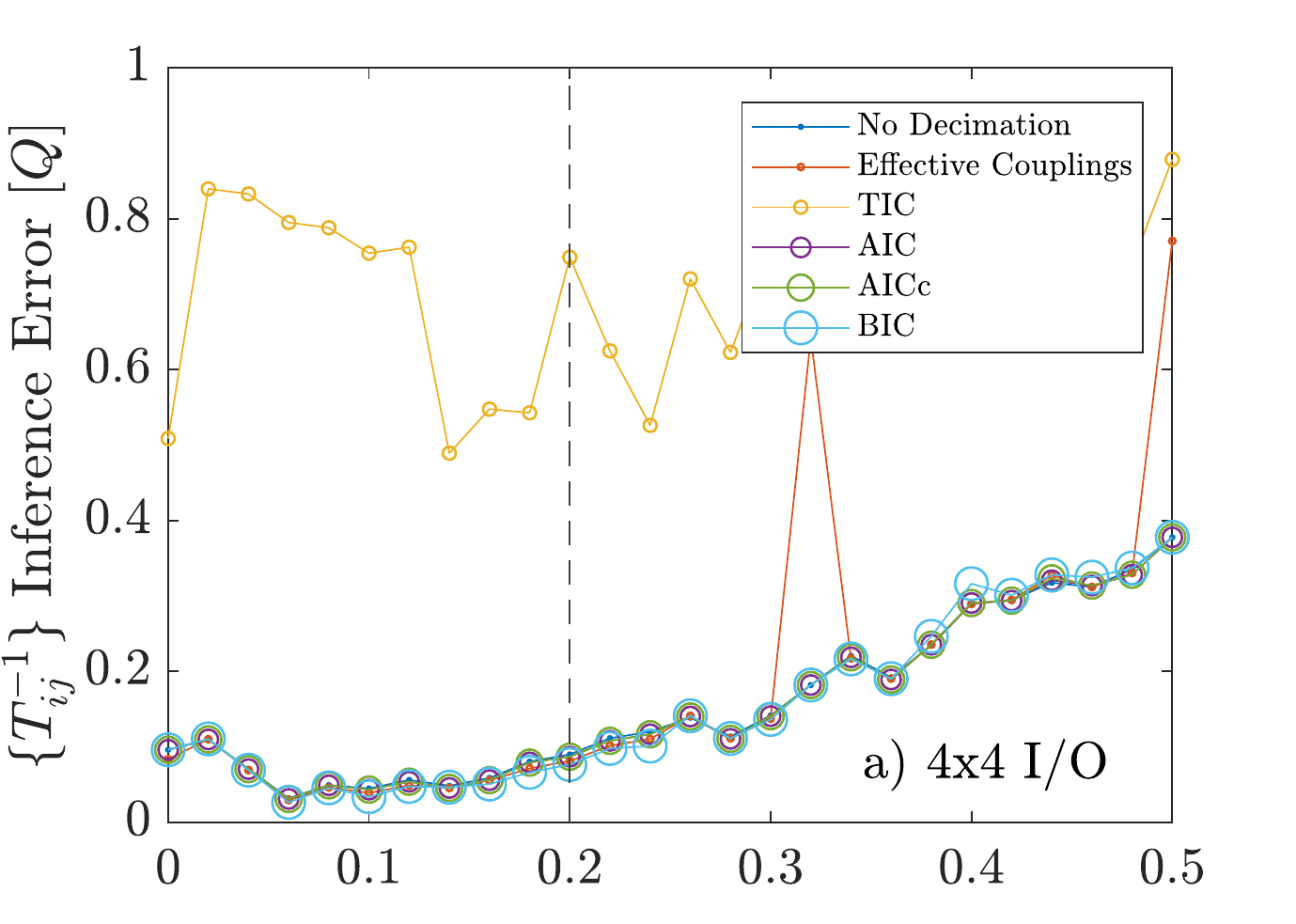}
\includegraphics[width=7cm]{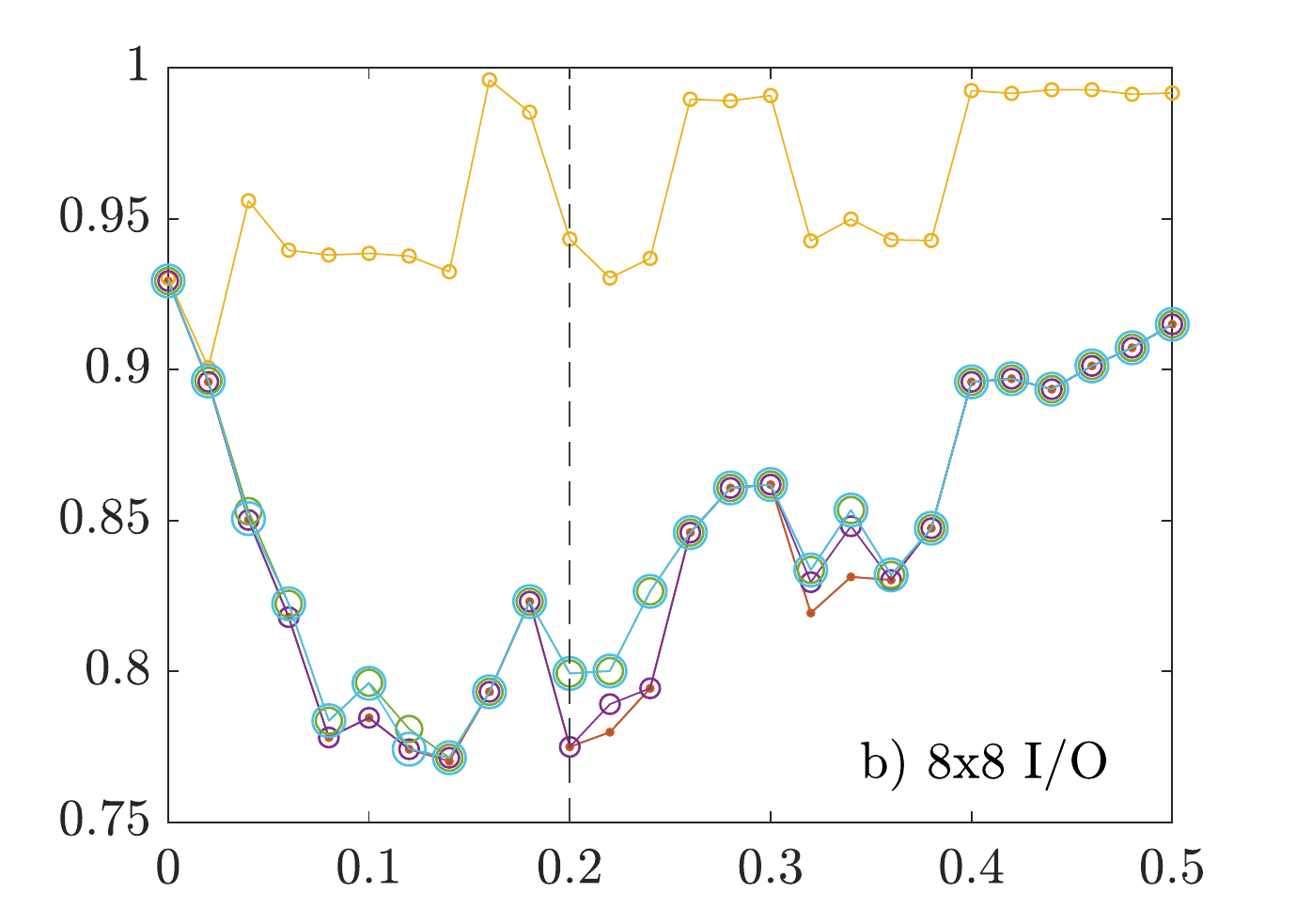}}
\centerline{\includegraphics[width=7cm]{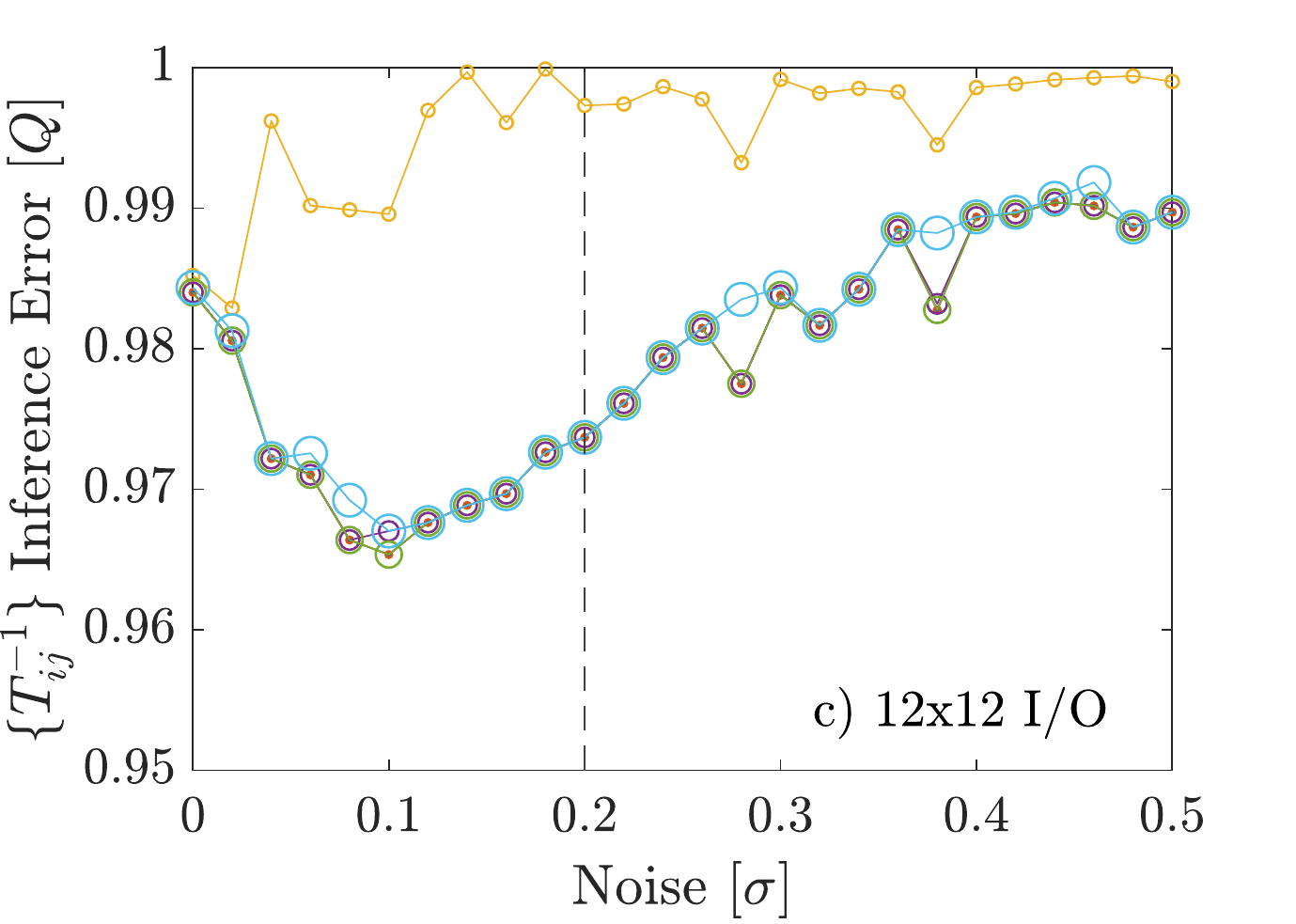}
\includegraphics[width=7cm]{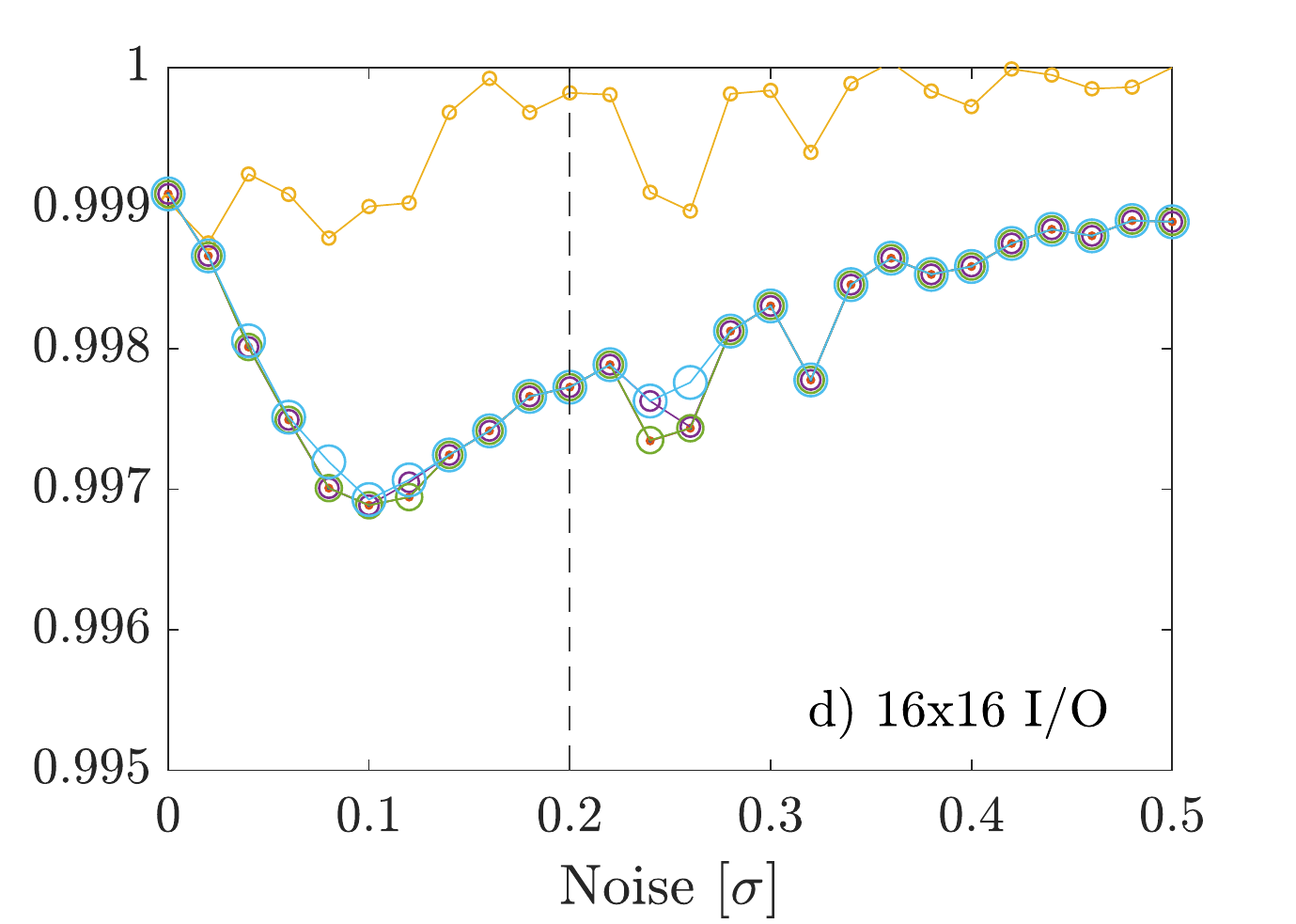}}
\caption{Reconstruction error of the inverse transmission matrix $\mathbb{T}^{-1}_{\rm inf}$. We plot the quantity $Q$ under different inference scenarios. Contrarily to Fig. \ref{fig:T_reconstruction_quality}, the curves do not start from zero, even if for the smallest system the error is quite low and grows with the noise. In all the other cases, the error is always quite high, having a minimum around $\sigma=0.1$.}
\label{fig:Tinv_reconstruction_quality}
\end{figure}

Interestingly, for the $\mathbb{T}^{-1}$ inference TIC always produced wrong estimation, strongly decimating coupling while the matrix was not sparse. 
This is more a test on TIC (and other IC, in general), then on the inference quality itself. Indeed, being $\mathbb{T}^{-1}$
a dense matrix we know in advance that any decimation trial should be irrelevant. In principle, no elements are zero, though many can be small, because of the stochasticity property $\sum_j  T^{-1}_{ij}=1$  (we recall that $\mathbb{T}$ is a stochastic matrix, with random entries $T_{ij} \in[0,1]$ and $\sum_j T_{ij}=1$). So decimating setting the smallest ones to zero, or leaving them undecimated, does not, actually, change the reconstruction outcome.

Yet, we have to take into account several issues, some strictly numerical. In a certain sense, the matrix $\mathbb{T}^{-1}$ was not directly controlled nor normalized (in the direct case we had full control on the definition of $\mathbb{T}$), and to estimate it we have to rely on numerical inversion techniques. Moreover, $\mathbb{T}^{-1}$ has also access to negative values that vary along each line, and the condition $\sum_j  T^{-1}_{ij}=1$ is not constraining the magnitude of the entries that can become large if proper cancellations occur. In these terms, the inference of direct- and inverse-$\mathbb{T}$ represent two opposite scenarios.

To control the inference quality in the inverse regime, we compute the reconstruction error $Q\left(\mathbb{T}^{-1},\left(\mathbb{T}^{-1}\right)_{\rm inf} \right)$ like in section \ref{sec:directT}. Compared to the situation described previously, we never reach $Q\approx0$ in any reconstruction. Moreover, seems that there is an optimal noise around $\sigma \approx 0.1$ at which the inference is slightly more accurate (minimum in the $Q$ curve). Again, all the IC selections tightly follow the same quality-trend being almost indistinguishable, with TIC quickly departing from good estimations already at $\sigma = 0.04$. By looking at both Fig. \ref{fig:inv_PLsurf} and \ref{fig:Tinv_reconstruction_quality}, it seems clear that TIC is not reliable while inferring $\mathbb{T}^{-1}$. TIC in fact, grounds its estimation based on sensitive features of the 
pseudo-likelihood  curve: a net drop of the $\mathcal{L}$ is picked easily with TIC, while noisy curves could missplace its maximum. Moreover, by its definition, the TIC cannot be used if the matrix is densely populated. TIC is defined to be zero at full or empty matrix connectivity, thus its maxima cannot be located at its edges.

\subsection{Pseudo-unity product}
\begin{figure}[t!!]
\centerline{\includegraphics[height=10cm]{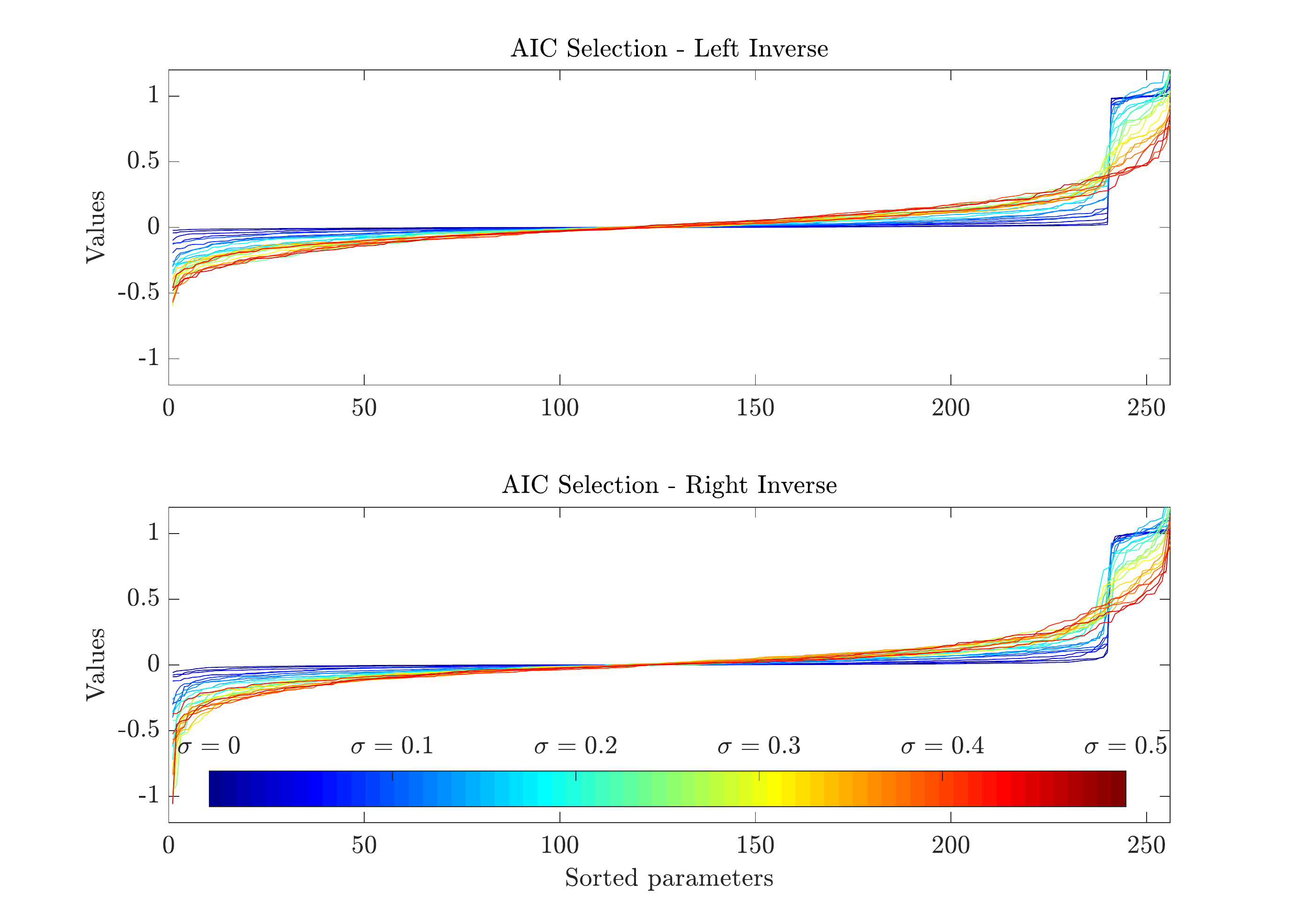}}
\caption{Sorted values of the matrix $\left(\mathbb{T}^{-1}\right)_{\rm inf} \mathbb{T}_{\rm inf}$ (top) and $\mathbb{T}_{\rm inf}\left(\mathbb{T}^{-1}\right)_{\rm inf}$
  (bottom) obtained with AIC selection with a $4$x$4$ system. The baseline of the identity matrix is drawn with a black solid line, while the sorted values at increasing noise are plotted following a progressive colorcode. The bluish lines (low $\sigma$) are closely following the expected behaviour, having a strong diagonal component dominating a almost null out of diagonal region. The situation progressively worsens at increased noise, though diagonal elements remain larger than the off-diagonal ones. }
\label{fig:unitaryproduct}
\end{figure}

Consistency is important especially when evaluating both $\mathbb{T}$ and its inverse independently. Thus, it is worth to study the product $\mathbb{T}^{-1} \mathbb{T}= \mathbb{T}\mathbb{T}^{-1}=\mathbb{I}$. 
The inference procedures are performed separately for each of the two matrices, consequently we are not guaranteed that this property is preserved during the inference. Moreover, we used different IC selections, that might perform better on direct or inverse inference but not with both, since $\mathbb{T}$
is sparse but $\mathbb{T}^{-1}$ is dense.
We tested the consistency of the method picking the best inferred matrices (direct and inverse) and multiplying each other, in order to estimate the identity matrix. Per each $w$ we use the information criterion that best matched the right number of couplings of the direct matrix, without any consideration for the inverse. In the plot of Fig. \ref{fig:unitaryproduct} we plot the sorted values of the pseudo-unity matrix $\left(\mathbb{T}^{-1} \right)_{\rm inf}\mathbb{T}_{\rm inf}$ calculated in the $w=4$ case, choosing AIC as the best selection criterion for both direct and inverse.

We use a progressive colormap to differentiate between reconstructions at different noise. In low noise transmission (bluish curves), the diagonal strongly dominates over the out-of-diagonal region, exibiting a visible step that tends to vanish as the rumour increases. However, we found that a dominant diagonal feature emerged at all the noises considered. For all the other size explored we found similar behaviour, with the general trend that the higher the number of measurements $M$ the better is the reconstruction of the diagonal part of the pseudo-unity matrix. As already discussed in section \ref{sec:inverseT}, TIC cannot be used to pick inverse matrices due to the fact that it is not devised for dense matrices.

\subsection{Testing image transmission and focusing}
The ultimate goal of this study, is to create a statistical framework able to recover the transmission matrix of a optical disordered system via a random sampling approach, opening the possibility to use turbid media as optically opaque lenses. Ideally, one wants a reliable system, able to correctly infer the properties of the channel on both directions even in the presence of strong noise fluctuation and saturation effects. To test the robustness of the inference procedure, we decided to accomplish two tasks defined in this way: 

\begin{list}{•}{}
\item \textbf{Focusing}: using the inferred $\mathbb T_{\rm inf }$ matrix to deliver energy to a given spatial pattern at the output. From the knowledge of $\mathbb T_{\rm inf}$  we are able to construct a given output pattern (like focusing on a given spot) changing the input. 

\item \textbf{Imaging}: using the inferred $\left(\mathbb{T}^{-1} \right)_{\rm inf}$ matrix to reconstruct the image of an object at the input starting from the noisy and random pattern in output.

\end{list}

In a perfect scenario, it might be reasonable to believe that both situations are equivalent  reversing the input-output order in the vector containing the pixel intensities  $\bm I$, due to the  exchange of role between $\mathbb T$ and $\mathbb T^{-1}$. 

To test the inferred matrices we take a further validation dataset $\bm I_{\rm val}$ composed of $M'=1000$ new patterns that we transmit through the disordered  channel. This important test acts as a validation for the training procedure, given that none of these $\bm I_{\rm val}$ vectors  were used in the learning dataset. By using $\mathbb T_{\rm inf}$ and $\left[\left(\mathbb T^{-1}\right)_{\rm inf}\right]^{-1}$ we want to reconstruct the intensity pattern focused at the output, $\bm I^{out}_{\rm rec}$, given an input.
 On the other hand by using $\left(\mathbb T^{-1}\right)_{\rm inf}$ and $ \left( \mathbb T_{\rm inf}\right)^{-1}$ we want to reconstruct the image at the input, $\bm I^{in}_{\rm rec}$, given a pattern at the output. To test the goodness of the reconstructions, we calculate the connected correlation $C$ between the true $\bm I^x$ and the reconstructed object $\bm I^x_{\rm rec}$, being $x={\rm in, out}$:
\begin{equation}
C \left(\bm I^x, \bm I^x_{\rm rec} \right) = \frac{ \sum_l \left( \bm I^x - \bar{\bm{I}}^x \right) \left( \bm I^x_{\rm rec} - \bar{\bm{I}}^x_{\rm rec} \right) }{  \sqrt{ \left[\sum_l \left( \bm I^x - \bar{\bm{I}}^x \right)^2 \right] \left[\sum_l \left( \bm I^x_{\rm rec} - \bar{\bm{I}}^x_{\rm rec} \right)^2 \right] } }
\end{equation}
where $l=\{\alpha, \gamma\}$ and $\bar{\bm{I}}^x$ is the mean value of ${\bm{I}}^x$. $C=1$ means exact reconstruction whereas a small $C$ means failed reconstruction. In principle $C$ could also access negative values, being $C=-1$ a perfectly reconstructed object with opposite sign.

\begin{figure}[t!]
\centerline{\includegraphics[width=7cm]{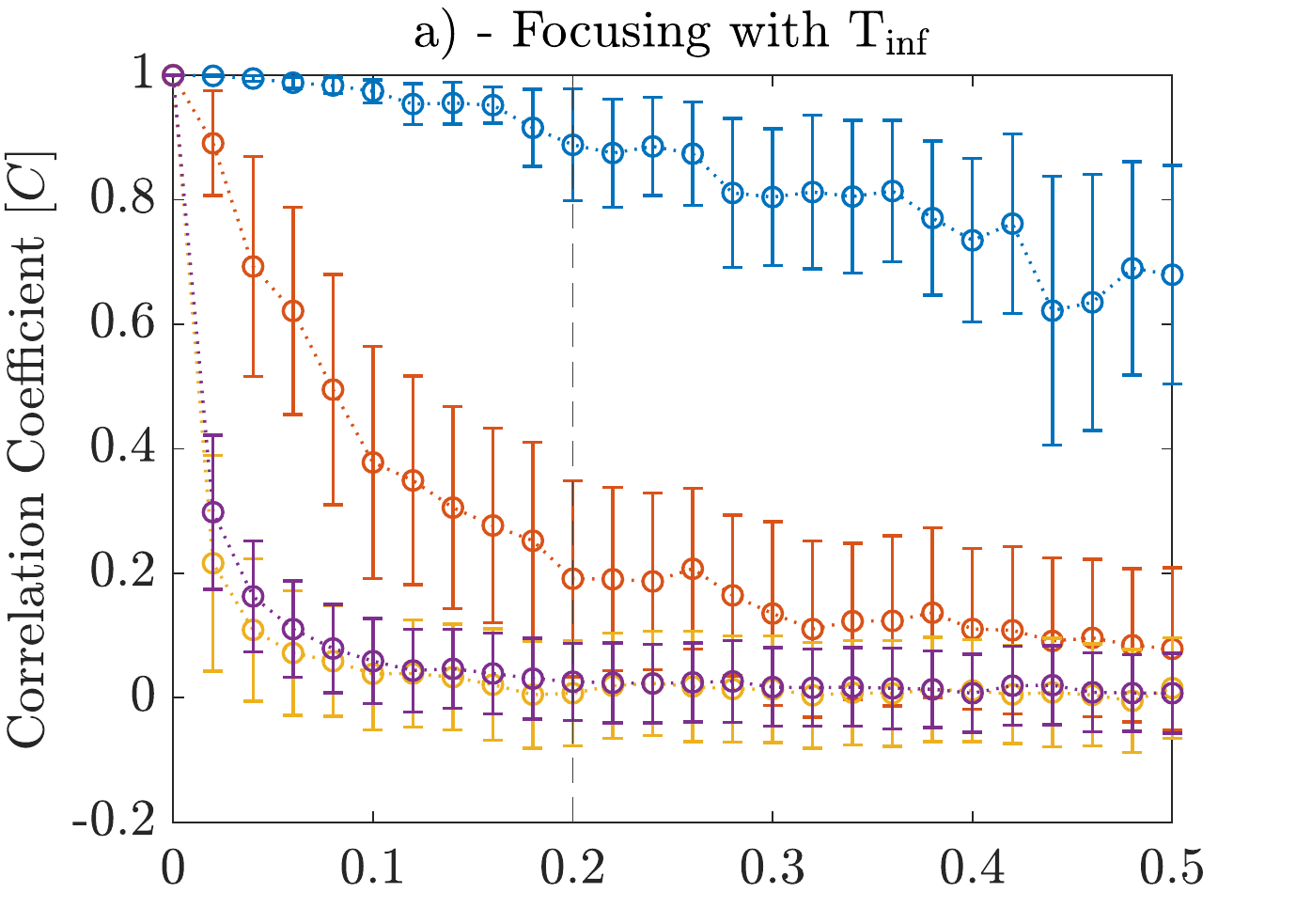}
\includegraphics[width=7cm]{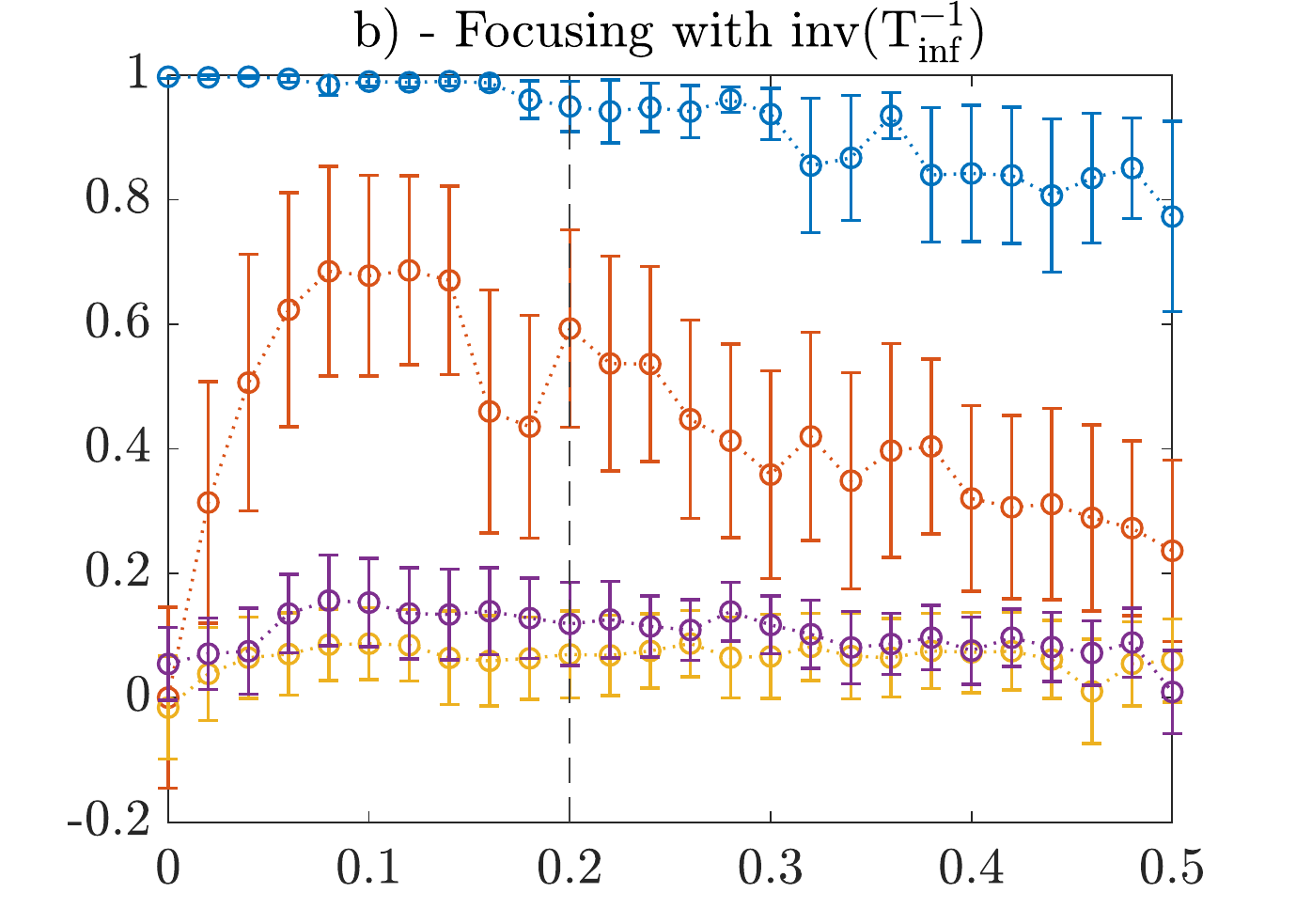}}
\centerline{\includegraphics[width=7cm]{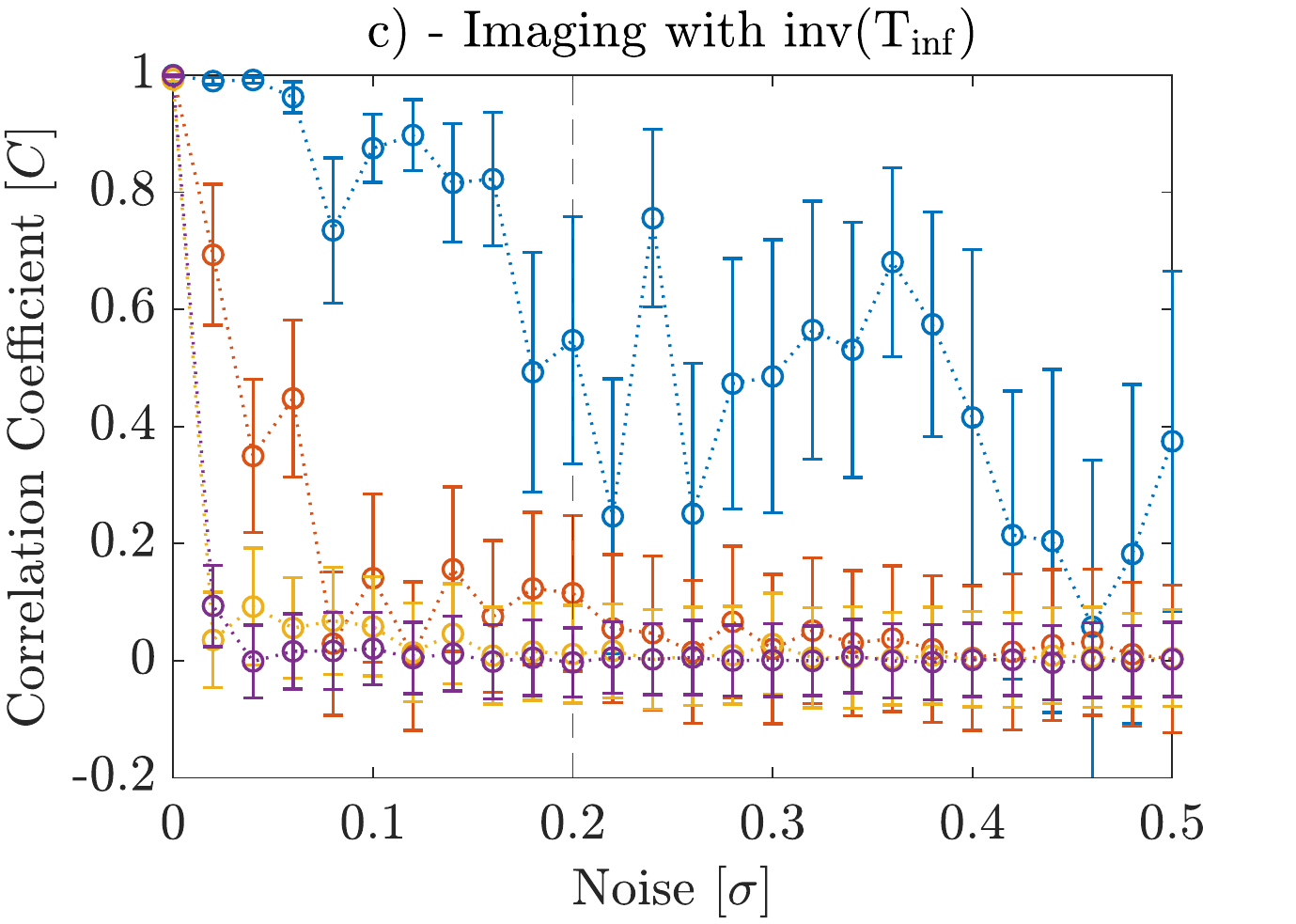}
\includegraphics[width=7cm]{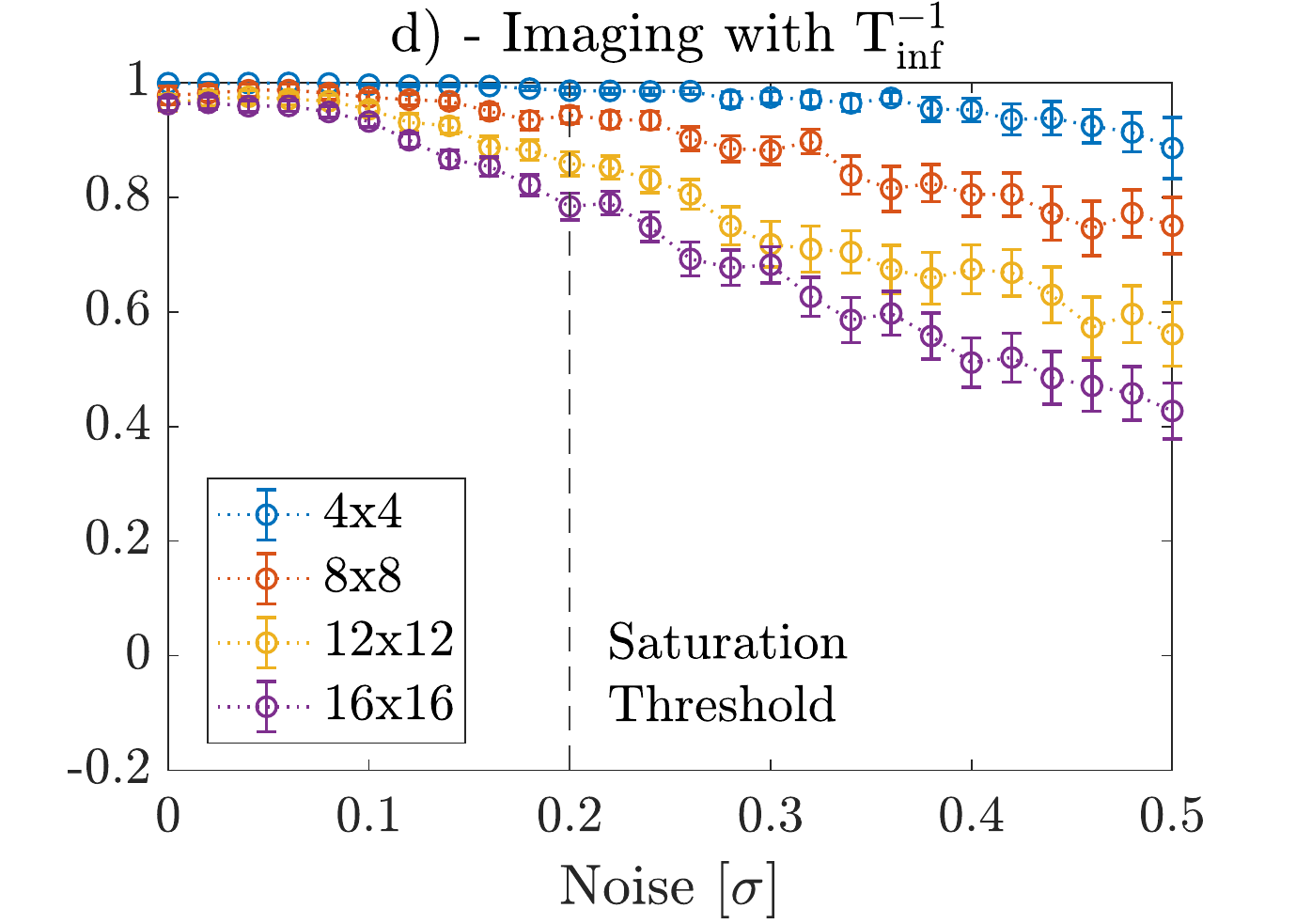}}
\caption{Correlation of the reconstructed patterns using the inferred $\mathbb T$ and $\mathbb T^{-1}$ and their corresponding inverted quantities.}
\label{fig:imaging_focusing}
\end{figure}

We report the analysis of the correlation behavior for focusing and for imaging versus sampling rate and noise in  Fig. \ref{fig:imaging_focusing}, where the transmission matrices used were inferred by pseudo-likelihood-maximization and AIC decimation.
\subsubsection*{Focusing}
\begin{enumerate}[label=\alph*)]
    \item Focusing with $\mathbb T_{\rm inf}$: it works extremely well at zero noise for any system linear size $w$. The correlation drops as soon as we increase the noise, 
    decreasing the faster, the smaller the sampling rate. For $\xi<1$ the focusing is substantially lost as soon as  $\sigma>0$, whereas for $\xi \gg1 $ the focusing with the real and the inferred matrices yield very correlated results also with a large noise $\sigma=0.5$.
    \item Focusing with $\left[\left( \mathbb T^{-1}\right)_{\rm inf}\right]^{-1}$: the inversion of the $\mathbb T^{-1}_{\rm inf}$  appears to be robust against inversion at high sampling rate for any noise, quickly dropping as $\xi \geq 1$. In these cases the correlation behavior is not monotonous with the noise but there seems to be an optimal noise value  ($\sigma \approx 0.1$) at which the inversion returns the best result.
\end{enumerate}

\subsubsection*{Imaging}
\begin{enumerate}[label=\alph*)]
  \setcounter{enumi}{2}
    \item Imaging with $\left( \mathbb T_{\rm inf}\right)^{-1}$: one of the main point of our study lies here. In fact, the numerical inversion of $\mathbb T_{\rm inf}$ is highly unstable already at moderate noise for practically any sampling rate $\xi$.  At $\sigma=0$, however, numerical inversion was correctly outperformed at all  investigated sizes, even for $\xi \ll 1$.
    \item Imaging with $\left(\mathbb T^{-1}\right)_{\rm inf}$: quite impressive results are found for the inverse inference. The correlation with the true pattern remains quite stable in the whole noise range explored, moreover being robust at any sampling scenario. \end{enumerate}

At this point of our study the situation is quite clear, matrix inversion is something that we need to avoid, especially in the presence of the noise. Both the matrix inversions resulted into less (or not-at-all) robust reconstruction, thus is always better to directly infer the  matrix needed for the given application: focusing or imaging . At the simple cost of doubling the inference time we have access to full characterization of the disordered system, more robust against noise. The problem  still scales linearly with the size of the system and it is still  possible to perform independent parallelization.

\section{Conclusions and Perspectives}
\begin{figure}[t]
\centerline{\includegraphics[width=14cm]{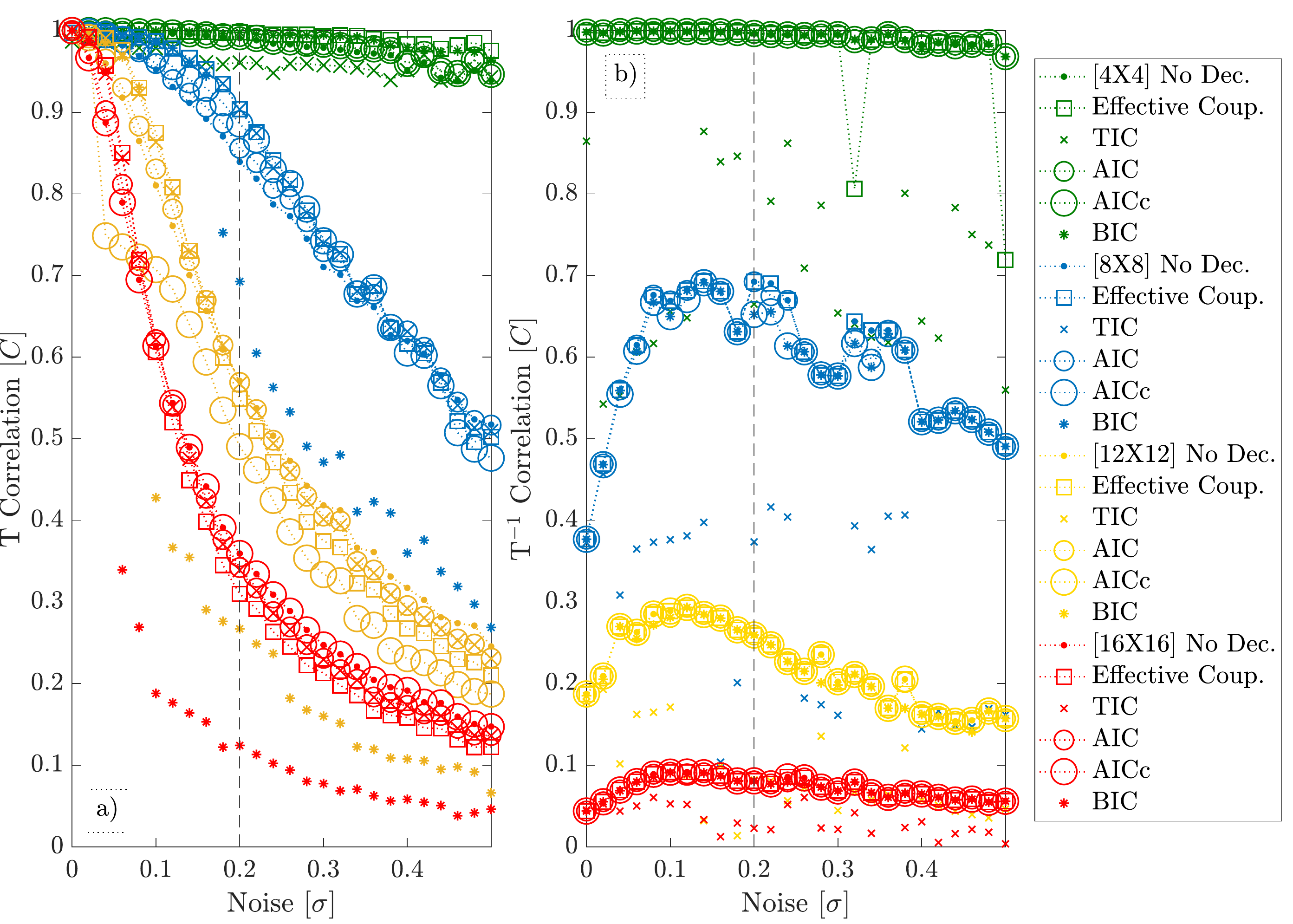}}
\caption{Part a), correlation of the $\mathbb{T}_{\rm inf}$ with the original one used to create both training and validation dataset as a function of the noise and for the various IC used. We can notice how every curve start with $C\approx1$ decaying faster as the system size increases.
Part b), $\mathbb{T}_{\rm inf}^{-1}$ correlation against the expected one. Here we notice a particularly low correlation, especially in the bigger (and undersampled) systems.}
\label{fig:dirinvT_correlation}
\end{figure}

At the end of such a long journey, we want to briefly discuss and recap some important aspects emerged from our study. First of all, our study wants to point out that relying to matrix inversion after the inference of $\mathbb{T}$ is a practice that is better to avoid in presence of non-negligible noise and small data sets (or large number of parameters to learn). Our model is equivalently reversible and accepts I/O swap, giving access to direct inference of the $\mathbb{T}^{-1}$. 
As a last additional analysis, let us examine the correlation coefficient of the $C(\mathbb{T}, \mathbb{T}_{\rm inf})$ and $C(\mathbb{T}^{-1}, \mathbb{T}_{\rm inf}^{-1})$. The results are presented in Fig. \ref{fig:dirinvT_correlation} as a function of the noise and the IC used for selecting among decimated models.

In panel a) the results of the correlation of the inferred transmission matrix against the true matrix are intuitive. The green curves prove better results when sampling $\xi \gg 1$, returning excellent correlation in the whole noise range explored. The situation progressively worsens as soon as $\xi$ decreases below $1$, i. e.,  the number of parameter becomes larger than the measurements used for the learning. 
\begin{figure}[t!]
\centerline{\includegraphics[width=14cm]{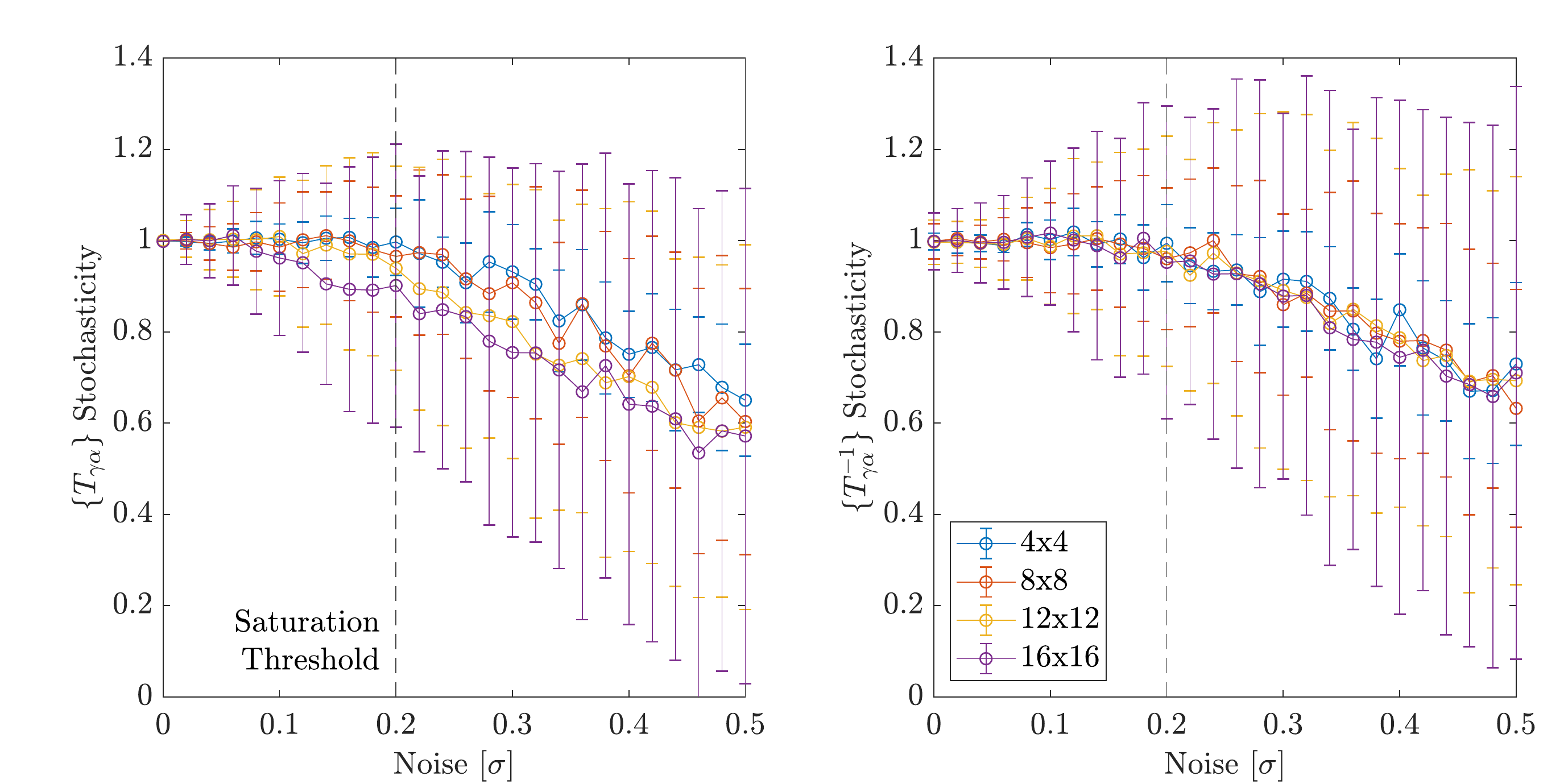}}
\caption{Measure of the stochasticity of the inferred matrix. We can notice that at zero noise is fully recovered in both cases, while increasing the noise the stochasticity decreases with huge fluctuations along lines.}
\label{fig:stochasticity}
\end{figure}
 On the other hand, we could look at the correlation trend of the $\mathbb{T}_{\rm inf}^{-1}$, reported in part b). It is standing out a pretty different situation. 
 Only the $\xi \gg 1$ case behaved like in the previous inference, where the others start from lower correlation at low noise. 
Afterwards, the correlation reaches a maximum then decerases again. The same behavior can be observed for any size, though for the two larger sizes 
of $12$x$12$ and $16$x$16$, the correlation with the original inverse matrix is so low that we may suspect poor reconstruction performances.
 
Since  the $\mathbb T$, and consequently $\mathbb T^{-1}$, is stochastic, it is interesting to observe if this property is recovered by the inference process, in which stochasticity is not imposed. Per each AIC reconstruction, we calculate the average row sum of $\mathbb T_{\rm inf}$ and $\left(\mathbb T_{\rm inf}\right)^{-1}$ that we plot against the noise in Fig. \ref{fig:stochasticity}. It is visible that the matrices recovered at any size and at low noises are stochastic. However, above the saturation threshold, the stochasticity starts to have strong fluctuations and its mean value decreases considerably. It is remarkable that the trend followed at every system size seems to coincide regardless of the sampling ratio.
 
\begin{figure}[h]
\centerline{\includegraphics[width=16cm]{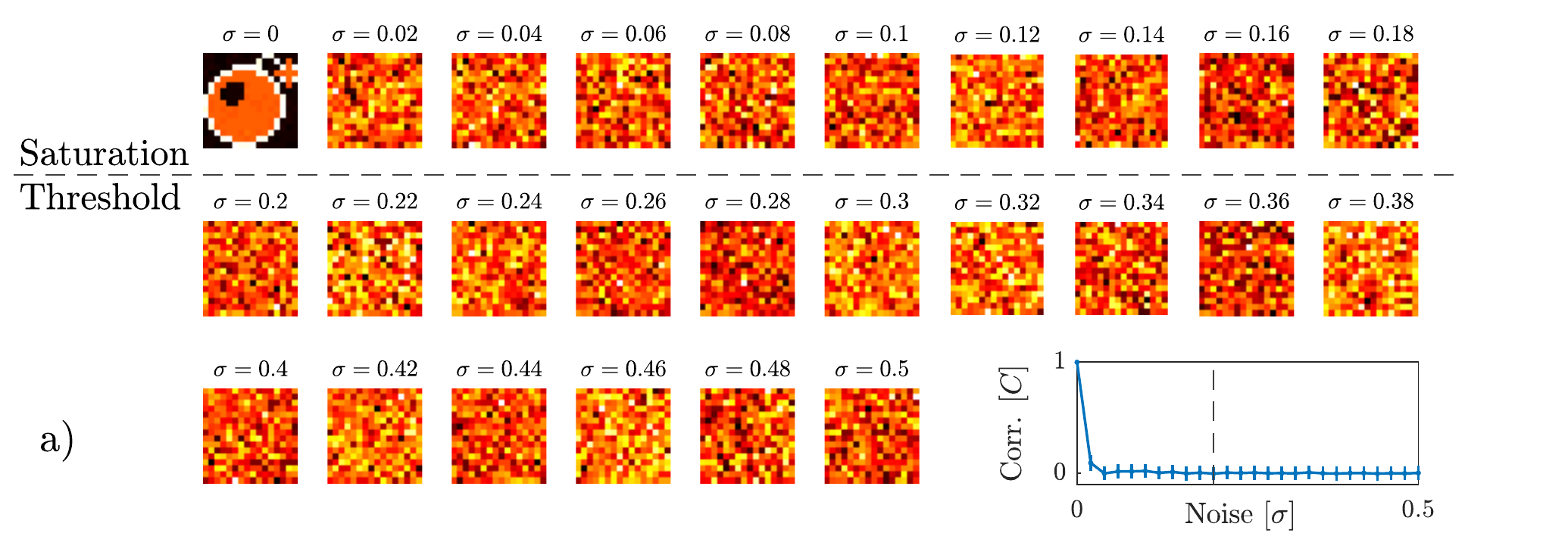}}
\centerline{\includegraphics[width=16cm]{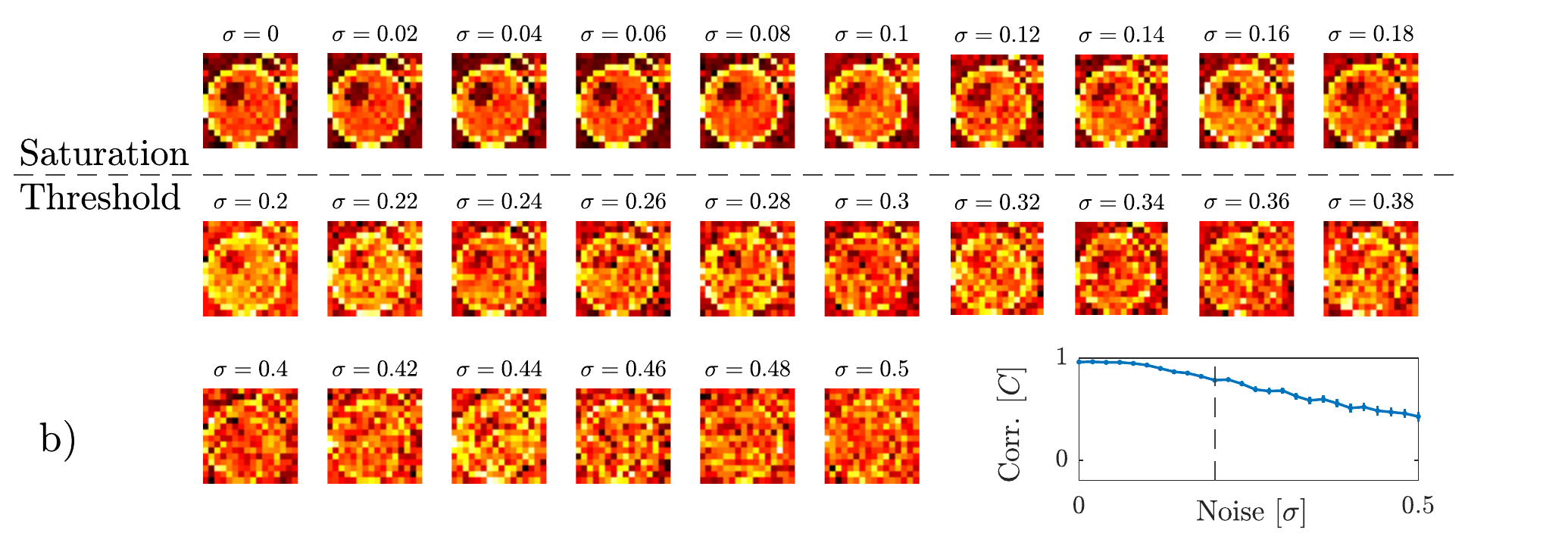}}
\caption{Upper part a), image recostructions using $\left( \mathbb{T}_{\rm inf}\right)^{-1}$ and bottom part b), using $\left(\mathbb{T}^{-1}\right)_{\rm inf}$. The insets reproduce the correlation between the image reconstructed with the original and with the inferred transmission matrices, cf. the $16\times 16$ curves of panel c (top) and d (bottom) of Fig. \ref{fig:imaging_focusing}. It is possible to notice how the performance are better in the inverse inference at any noise, where the direct gives excellent reconstruction only at zero noise.}
\label{fig:bomb_reconstruction}
\end{figure}

The correlation plot of Fig. \ref{fig:dirinvT_correlation} seems to condemn the inference of $\mathbb T^{-1}$ to low quality reconstructions in the case of low sampling rate. On the contrary, though, we tested such matrices obtaining very robust reconstructions for the validation set in Fig. \ref{fig:imaging_focusing}. 
Let us, then, have a look at an image reconstructed using $\left(\mathbb{T}^{-1}\right)_{\rm inf}$ and using $\left(\mathbb{T}_{\rm inf}\right)^{-1}$ (also termed inv($\mathbb{T}_{\rm inf})$ in the figures). As a test image we use a 16x16 image of a little bomb, reconstructing the image propagating from the output back to the input edge. 
The reconstructions are presented in Fig. \ref{fig:bomb_reconstruction} using $( \mathbb{T}_{\rm inf})^{-1}$ (set on top) and $\left(\mathbb{T}^{-1}\right)_{\rm inf}$ (bottom). In the first case, the reconstruction is perfect at zero noise, but its quality immediately drops and the image is not recognizable at any noise other than $\sigma=0$. The latter, instead, returns remarkable reconstructions at finite noise, even if the $\mathbb{T}^{-1}$-correlation plot was poor and the reconstruction error $Q$ was large. At noises around the saturation threshold $\sigma \neq 0.2$ the reconstructions start to fade into noisy background.

At this point, we can sum up our take home message:
 the inversion process carried on a matrix inferred by pseudo-likelihood maximization on undersampled training dataset leads to poor results, as expected. 
In our framework, though, the inference of the inverse seems performing well even when datasets are small and offers an extra tool exploitable for focusing or imaging through disordered channels.  The inference procedure on the inverse transmission matrix for low sampling ratio's yields a matrix that does not reproduce well the original one, yet (i) gives the identity matrix when multiplied to the direct $\mathbb{T}$ and (ii) performs high quality image reconstruction when used on validation data sets.

Although the model Eq. (\ref{eq:transmission}) is not rigorously correct for coherent field summation, it is a first approximation necessary to deal with intensity only data and a first step to consider the inaccessibility of mode phases. This article, in fact, has to be considered as a complete treatise on the method firstly introduced in \cite{ancora2019learning}. The vastness of the results found pushed us toward a full characterization of the learning framework introduced.
As work in progress we are developing an algorithm that takes into account the impossibility to access to phase measurements both at input and output edges \cite{Ancora20}. Our approach is to calculate the $\mathcal{L}$ performing first a weighted integral over the unknown phases with a Boltzmann equilibrium measure, i. e., under the same hypothesis of Eq. (\ref{eq:probability}).  In this way, we obtain an expression of $\mathcal{L}$ usable for intensity measurements in which, at the same time, the parameters to be inferred are the elements of the actual transmission matrix (i. e., part of the scattering matrix) of the object crossed by the light, and
 not  an effective matrix connecting I/O pixels. 
 Therefore, this will also allow to study properties of the structure of the materials composing the medium, on top of yielding a tool for focusing and/or image reconstruction.

Among conventional focusing and imaging through disordered media \cite{vellekoop2010exploiting, Popoff2010natcomm}, our approach finds a major application for the comprehension of structured focusing \cite{DiBattista2016, di2016tailored}, to study Anderson localization in disordered fibers \cite{Leonetti2014} or signal transmission through hyperuniform channels \cite{di2018hyperuniformity}. 
Moreover, the bidirectional approach let our framework consistent for the study of truncated channels: when observing a fiber, we can look at the whole fiber's output compared with the whole input, while this is not always true in case the observation in done in a confined output region \cite{Popoff2010natcomm}. Leaving open channels in the system leads the impossibility of $\mathbb T$-matrix inversion, and forces us to infer directly its inverse.

Even though we used as example the case of the transmission matrix recovery in a disordered optical tool, our results are generally extendible for any linear system solution and has a broad application spectrum, especially in the context of supervised machine learning \cite{robert2014machine, friedman2001elements}. 
A particular feature of our model is that a lot of information is encoded in the interaction matrix $\mathbb M$ other than the transmission matrix alone. 
In fact, we are automatically obtaining an estimation of the noise variance under Gaussian statistics (thus giving estimation on the presence of more noisy channels) and the self-input coupling term is proportional to the Gramian matrix that can be exploited for further analysis \cite{sreeram1994properties}. 

From a technical point of view, one important feature is the linear scalability in the model complexity as function of the dimension of the system considered. Each partial likelihood term $\mathcal{L}_i$ is an independent entity and their summation can always be parallelized among the hardware resources. Our model, in fact, finely fits the GPU requirements and it is directly parallelizable in CUDA. 
Last, but not least, this work was inspired by recent development in statistical mechanics of random optical systems \cite{Antenucci15b,Antenucci15e,Antenucci16}, either from thermodynamics and from inference work \cite{Tyagi2016, Marruzzo2017a}. From this world we borrowed a number of ideas \cite{Tyagi2016, Decelle2014} and further tools could be linked to expand the results to further cases, studying for example phase transitions between feasible and unfeasible inference scenarios.

\section*{Acknowledgements}
The authors thank  A. Marruzzo,  G. Gradenigo, F. Antenucci, J. Rocchi, F. Ricci-Tersenghi, L. Biferale, for useful discussions. 



\paragraph{Funding information}
The research leading to these results has received funding from the Italian Ministry of Education, University and Research under the PRIN2015 program, grant code 2015K7KK8L-005 and the European Research Council (ERC) under the European Union's Horizon 2020 research and innovation program, project LoTGlasSy, Grant Agreement No. 694925.

\bibliography{bibliography_SciPost}

\nolinenumbers

\end{document}